\documentclass{article}

\usepackage{microtype}
\usepackage{graphicx}
\usepackage{booktabs} 

\usepackage{hyperref}

\usepackage[accepted]{icml2023}

\usepackage{amsmath}
\usepackage{amssymb}
\usepackage{mathtools}
\usepackage{amsthm}

\usepackage{csquotes}
\usepackage{caption}
\usepackage{subcaption}
\usepackage{bm}
\usepackage{pifont}

\newcommand{\defeq}{\doteq}
\newcommand{\subX}{{\bm {\mathsf{X}}}}
\newcommand{\R}{\mathbb{R}}

\newcommand{\shead}{\text{SHead}}
\newcommand{\sattn}{\text{SAttn}}
\newcommand{\stb}{\text{STB}}

\DeclareMathOperator*{\argmax}{arg\,max}

\usepackage[capitalize,noabbrev]{cleveref}

\theoremstyle{plain}
\newtheorem{theorem}{Theorem}[section]

\newtheorem{lemma}[theorem]{Lemma}
\newtheorem{corollary}[theorem]{Corollary}
\theoremstyle{definition}
\newtheorem{definition}[theorem]{Definition}

\theoremstyle{remark}

\usepackage[textsize=tiny]{todonotes}

\icmltitlerunning{Sampled Transformer for Point Sets}

\begin{document}

\twocolumn[
\icmltitle{Sampled Transformer for Point Sets}




\begin{icmlauthorlist}
\icmlauthor{Shidi Li}{yyy}
\icmlauthor{Christian Walder}{comp}
\icmlauthor{Alexander Soen}{yyy}
\icmlauthor{Lexing Xie}{yyy}
\icmlauthor{Miaomiao Liu}{yyy}
\end{icmlauthorlist}

\icmlaffiliation{yyy}{Australian National University, Canberra, Australia}
\icmlaffiliation{comp}{Google Brain, Montreal, Canada}

\icmlcorrespondingauthor{Shidi Li}{shidi.li@anu.edu.au}

\icmlkeywords{Machine Learning, Transformer, 3D Point Cloud}

\vskip 0.3in
]

\printAffiliationsAndNotice{} 

\begin{abstract}
    The sparse transformer can reduce the computational complexity of the self-attention layers to $O(n)$, whilst still being a universal approximator of continuous sequence-to-sequence functions.
    However, this permutation variant operation is not appropriate for direct application to sets.
    In this paper, we proposed an $O(n)$ complexity sampled transformer that can process point set elements directly without any additional inductive bias.
    Our sampled transformer introduces random element sampling, which randomly splits point sets into subsets, followed by applying a shared Hamiltonian self-attention mechanism to each subset.
    The overall attention mechanism can be viewed as a Hamiltonian cycle in the complete attention graph, and the permutation of point set elements is equivalent to randomly sampling Hamiltonian cycles.
    This mechanism implements a Monte Carlo simulation of the $O(n^2)$ dense attention connections. 
   We show that it is a universal approximator for continuous set-to-set functions. 
  Experimental results on point-clouds show comparable or better accuracy with significantly reduced computational complexity compared to the dense transformer or alternative sparse attention schemes. 
\end{abstract}

\section{Introduction}
Encoding structured data has become a focal point of modern machine learning. In recent years, the defacto choice has been to use transformer architectures for sequence data, \emph{e.g.}, in language~\citep{vaswani2017attention} and image~\citep{dosovitskiy2020image} processing pipelines.
Indeed, transformers have not only shown strong empirical results, but also have been proven to be universal approximators for sequence-to-sequence functions~\citep{yun2019transformers}.
%
%
%
%
Although the standard transformer is a natural choice for set data due to permutation invariant dense attention, its versatility is limited by the costly \(O(n^2)\) computational complexity.
To decrease the cost, a common trick is to use sparse attention, which reduce the complexity from $O(n^2)$ to $O(n)$~\citep{guo2019star,yun2020n,zaheer2020big}.
However, in general this results in an attention mechanism that is not permutation invariant -- swapping two set elements change which elements they attend.
As a result, sparse attention cannot be directly used for set data.

Recent work has explored the representation power of transformers in point sets as a plug-in module~\citep{lee2019set}, a pretraining-finetuning pipeline~\citep{yu2022point,pang2022masked}, and with a hierarchical structure~\citep{zhao2021point}.
However, these set transformers introduced additional inductive biases to (theoretically) approach the same performance as the densely connected case in language and image processing applications.
Here inductive bias refers to the prior knowledge and design built into a machine learning model.
For example, to achieve permutation invariance with efficient computational complexity, previous work has required additional inductive bias such as nearest neighbor search~\citep{zhao2021point} or inducing points sampling~\citep{lee2019set}.
Detailed discussion could be found in \S \ref{app:inductive_bias} in the supplementary material.
Following the above analysis, a research question naturally arises to avoid introducing unneeded inductive bias:
%
\begin{displayquote}
\textit{Can $O(n)$ complexity sparse attention mechanisms be applied directly to sets?} 
\end{displayquote}

\begin{figure*}[ht]
    \centering
    \begin{subfigure}[b]{0.25\textwidth}
        \centering
        \includegraphics[width=\textwidth]{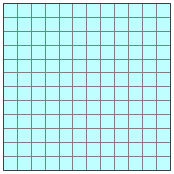}
        \caption{Dense Attention}
        \label{fig:dense}
    \end{subfigure}
    \hfill
    \begin{subfigure}[b]{0.25\textwidth}
        \centering
        \includegraphics[width=\textwidth]{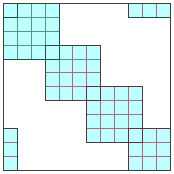}
        \caption{After Sampling}
        \label{fig:after_sampling}
    \end{subfigure}
    \hfill
    \begin{subfigure}[b]{0.25\textwidth}
        \centering
        \includegraphics[width=\textwidth]{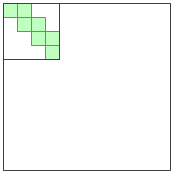}
        \caption{Hamiltonian Attention}
        \label{fig:Hamiltonian}
    \end{subfigure}
    \hfill
    \begin{subfigure}[b]{0.25\textwidth}
        \centering
        \includegraphics[width=\textwidth]{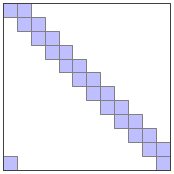}
        \caption{Cycle Attention}
        \label{fig:cycle}
    \end{subfigure}
    \hfill
    \begin{subfigure}[b]{0.25\textwidth}
        \centering
        \includegraphics[width=\textwidth]{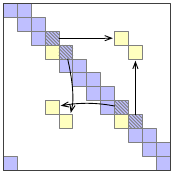}
        \caption{Swap Points}
        \label{fig:swap}
    \end{subfigure}%
    \hfill
    \begin{subfigure}[b]{0.25\textwidth}
        \centering
        \includegraphics[width=\textwidth]{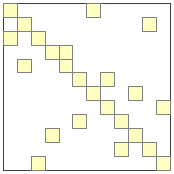}
        \caption{Sampled Attention}
        \label{fig:sampled}
    \end{subfigure}%
    \caption{
        Attention mechanisms:
        \textit{(a)} original \textit{dense attention};
        \textit{(b)} the attention matrix after random element sampling;
        \textit{(c)} a special case of sparse attention -- \textit{Hamiltonian (self-)attention} -- for each subset;
        \textit{(d)} combining all subsets (which have overlapping element per (b)) connects the individual Hamiltonian attention sub-matrices, gives \textit{cycle attention} which is a Hamiltonian cycle;
        \textit{(e)} permutation of points permutes the elements in cycle attention matrix;
        \textit{(f)} the resulting \textit{sampled attention}, viewed as a sampled Hamiltonian cycle from the edges of the complete attention graph.
    }
    \label{fig:diagram}
\end{figure*}

We propose the sampled transformer to address this question, which is distinguished from the original sparse transformer by mapping the permutation of set elements to the permutation of attention matrix elements.
Viewing this permutation sampling as attention matrix sampling, the proposed sampled attention approximates \( O(n^2) \) dense attention.
This is achieved with the proposed random element sampling and Hamiltonian self-attention.
To be specific, in random element sampling the input point set is first randomly split into several subsets of $n_s$ points (Fig.~\ref{fig:after_sampling}), each of which will be processed by shared self-attention layers.
In addition, a sparse attention mechanism -- namely \textit{Hamiltonian self-attention} (Fig.~\ref{fig:Hamiltonian}) -- is applied to reduce complexity of the subset inputs, so that $n_s$ point connections are sampled from $O(n_s^2)$ connections.
The combination of all Hamiltonian self-attention mechanism for all subsets -- namely \textit{cycle attention} (Fig.~\ref{fig:cycle}) -- can be viewed as a Hamiltonian cycle in the complete attention graph.
As a result, the permutation of set elements is equivalent to the permutation of nodes in a Hamiltonian cycle (Fig.~\ref{fig:swap}), which is in fact randomly sampling Hamiltonian cycles from the complete graph -- thereby yielding the proposed \textit{sampled attention} (Fig.~\ref{fig:sampled}).
Finally, viewing this randomization as a Monte Carlo sample of attention pairs, repeated sampling can be used to approximate the complete \( O(n^2) \) dense connections.
Furthermore, our proposed sampled transformer is proven to be a universal approximator for set data -- means any continuous set-to-set functions can be approximated to arbitrary precision.

The contributions of this paper are summarized as follows.
\begin{itemize}
    \item We propose the sampled attention mechanism which maps the random permutation of set elements to the random sampling of Hamiltonian cycle attention matrices, permitting the direct processing of point sets. 
    \item We prove that the proposed sampled transformer is a universal approximator of continuous set-to-set functions, see Corollary~\ref{thm:our_ua}.
    \item Compared to previous transformer architectures, the empirical results show that our proposed sampled transformer achieves comparable (or better) performance with less inductive bias and complexity.
\end{itemize}

\section{Related Work}
The transformer~\citep{vaswani2017attention} is widely used in languages~\citep{dai2019transformer, yang2019xlnet, raffel2020exploring} and images~\citep{ramachandran2019stand, dosovitskiy2020image, liu2021swin, touvron2021training}.
For example, \citet{raffel2020exploring} explored the transformer by unifying a suite of text problems to a text-to-text format;
\citet{dai2019transformer} modeled very long-term dependency by reusing previous hidden states;
\citet{dosovitskiy2020image} demonstrated that the pure transformer can be effectively applied directly to a sequence of image patches;
and \citet{liu2021swin} proposed a transformer with hierarchical structure to learn various scales with linear computational complexity.
In addition, the representation power of the transformer has been explored by the pre-training and fine-tuning models~\citep{bao2021beit, yu2022point, he2022masked}.

Recently, an increasing number of researchers begin to explore the representation power of the transformer in 3D point clouds (sets) data.
\citet{xie2018attentional} applied multi-layered dense transformers to small-scale point clouds directly;
\citet{yang2019modeling} further proposed the Group Shuffle attention to deal with size-varying inputs by furthest point sampling;
\citet{han2022dual} aggregated point-wise and channel-wise features by directly adding two self-attention layers.
To avoid the tricky tokenization step, \citet{lee2019set} tried to deal with points directly with $O(nm)$ complexity by introducing inducing points, and proved universal approximation;
\citet{mazur2021cloud} further proposed a hierarchical point set mapping, grouping, and merging structure with nearest neighbors defining the sparse attention mechanism.
\citet{yu2022point} and \citet{pang2022masked} further introduced the transformers to the pre-training and fine-tuning pipelines in the area of 3D point clouds.
Last but not the least, transformers have also been widely used in other such works on 3D (point cloud) data as \cite{liu2019point2sequence,fuchs2020se,misra2021end,mao2021voxel,sander2022sinkformers}

Another important line of work seeks to theoretically demonstrate the representation power of the transformer by showing the universal approximation of continuous sequence-to-sequence functions~\citep{yun2019transformers,yun2020n,zaheer2020big,shi2021sparsebert,kratsios2021universal}. 
To be specific, \citet{yun2019transformers} demonstrated the universal approximation property of the transformer;
\citet{yun2020n} and \citet{zaheer2020big} demonstrated that the transformer with sparse attention matrix remains a universal approximator;
\citet{shi2021sparsebert} claimed that the transformer without diag-attention is still a universal approximator.
\citet{kratsios2021universal} proposed that the universal approximation under constraints is possible for the transformer.

In comparison with the above works, we proposes the $O(n)$ \textit{sampled transformer} -- a universal approximator of continuous set-to-set functions. To our knowledge, the use of approximating dense attention by sampling Hamiltonian cycle attention matrices is new.

\section{Preliminary}

\subsection{Notation}
Given an integer \( a \) we define \( [a] \defeq \{ 1, \ldots, a \} \).
For a matrix \( {\bm M} \in \R^{n \times m} \), for a \(k \in [m] \) the \( k\)-th column is denoted by \( {\bm M}_{k} \). Given an (ordered) index set \( \mathcal{A} \subset [m] \) the submatrix \( {\bm M}_{\mathcal{A}} \in \R^{n \times \vert \mathcal{A} \vert} \) consists of the matrix generated by concatenating the columns determined by indices in \( \mathcal{A} \).
See the notation guide in \S\ref{app:notation} in the supplementary material.

\subsection{Transformer}\label{subsec:tf}
The transformer $\bm X \mapsto t(\bm X)$~\citep{vaswani2017attention,dosovitskiy2020image} implements a function from point clouds to point clouds with input points $\bm X \in \mathbb{R}^{d\times n}$.
It is formally defined by a multi-head self-attention layer and a feed-forward layer:
\begin{subequations}
\begin{align}
    \label{eq:dense_self_attn}
   &\text{Head}^j(\bm X) = (\bm W^j_V \bm X) \cdot \sigma_S [ (\bm W^j_K \bm X)^T \bm W^j_Q \bm X]\\
   &\text{Attn}(\bm X) = \bm X + \bm W_O \begin{bmatrix} \text{Head}^1(\bm X) \\ 
    \vdots \\ \text{Head}^h(\bm X) \end{bmatrix}\\
   &\text{TB}(\bm X) = \text{Attn}(\bm X) + \bm W_2 \cdot \text{ReLU}(\bm W_1 \text{Attn}(\bm X)),
\end{align}%
\label{eq:dense_tf}%
\end{subequations}%
where $n$ is the number of points and $d$ is the feature dimension.
$\text{Head}(\cdot)$ is the \textbf{self-attention layer}, and $\text{Attn}(\cdot)$ is the \textbf{multi-head self-attention layer} with the parameter $\bm W_O \in \mathbb{R}^{d\times mh}$.
$\bm W^i_V, \bm W^i_K, \bm W^i_Q \in \mathbb{R}^{m\times d}$ are value, key, and query parameters;
$\bm W_1 \in \mathbb{R}^{r\times d}$ and $\bm W_2 \in \mathbb{R}^{d\times r}$ are feed-forward layer parameters.
We utilize a positional embedding $\bm E$ in the input $\bm X$, defined by $\bm E = \bm W_p \bm P$, where $\bm P \in \mathbb{R}^{3\times n}$ is the ($xyz$) coordinate, and  $\bm W_P \in \mathbb{R}^{d\times 3}$ is an MLP layer.
To simplify the notation, here we use $\bm X = \bm X + \bm E$ so that all the inputs $\bm X$ in this paper will include the positional embedding unless specifically stated otherwise.
The \textbf{attention mechanism} for a dense transformer is the $n\times n$ attention matrix $(\bm W^i_K \bm X)^T \bm W^i_Q \bm X$ in Eq.~\ref{eq:dense_self_attn}, which is in fact a similarity matrix for $n$ elements/tokens, or a \textbf{complete attention graph}.
\textbf{Sparse Attention} also refers to the same similarity matrix/attention graph but with sparse connections instead.
As tokenization may not be necessary in dealing with point clouds, for clarity we use the terminology points, elements, and tokens are all to refer to points (which may be thought of as tokens in a traditional transformer context) in a point cloud (set).

\subsection{Universal Approximation}\label{subsec:ua}
Let $\mathcal{F}$ be the class of continuous sequence-to-sequence functions  $f: \mathbb{R}^{d\times n} \mapsto \mathbb{R}^{d\times n}$ defined on any compact domain.
Further define $\mathcal{T}^{h,m,r}$ as the set of transformer blocks $t(\cdot)$ with $h$ attention heads of each of size $m$, and with hidden layer width $r$~\citep{yun2019transformers,yun2020n}.
To measure the distance between functions in \( \mathcal{F} \), we define the standard \( \ell_{p} \) distance function by the corresponding norm:
\begin{align}
   d_p(f_1, f_2) = \left( \int \Vert f_1(\bm X) - f_2(\bm X) \Vert^p_p \; \mathrm{d}\bm X \right)^{1/p},
\end{align}
which is element-wise continuous (w.r.t the $\ell_p$ norm) for $1\leq p < \infty$.

\begin{theorem}[Universal Approximation,~\citet{yun2019transformers}]\label{theorem:ua}
   Let $1 \leq p < \infty$ and $\epsilon > 0$, then for any given $f\in \mathcal{F}$, there exist a Transformer network $g\in \mathcal{T}^{2,1,4}$, such that $d_p(f, g) \leq \epsilon$.
\end{theorem}

The proof of Theorem~\ref{theorem:ua} makes three stages of approximations, which are chained together via the triangle inequality to give the \( \epsilon \) bound \citep{yun2019transformers}. In particular, \ding{172} any \( f \in \mathcal{F} \) is approximated by a piece-wise linear function \( \overline{f} \in \overline{\mathcal{F}} \) (over a discretized input space). Then \ding{173} the piece-wise linear function is approximated by a \emph{modified} transformer \( \overline{\mathcal{T}}^{2,1,4} \), where the widely used \( \text{ReLU} \) and \( \sigma_{S} \) activation functions (as per Eq.~\ref{eq:dense_tf}) are replaced by the hardmax function \( \sigma_{H} \). Finally, \ding{174} it is shown that the class of transformer \( \overline{\mathcal{T}}^{2,1,4} \) can approximate any regular transformer \( g \in {\mathcal{T}}^{2,1,4} \).

The key step comes in the proof of the second approximation \ding{173}. In \citet{yun2019transformers}, the approximation is proved by showing that multi-head self-attention layers of the modified transformer can implement any contextual map \( q_c: \mathbb{R}^{d\times n} \mapsto \mathbb{R}^{n} \).
\begin{definition}[Contextual Mapping]\label{def:contextual_mapping}
   Consider a finite set $\mathbb{L} \subset \mathbb{R}^{d\times n}$. A map q: $\mathbb{L} \mapsto \mathbb{R}^{1\times n}$ defines a contextual map if the map satisfies the following:
   \begin{enumerate}
    \item For any $\bm L \in \mathbb{L}$, the $n$ entries in $q(\bm L)$ are all distinct.\label{def:contextual_mapping_rule1}
    \item For any $\bm L, \bm L' \in \mathbb{L}$, with $\bm L \neq \bm L'$, all entries of $q(\bm L)$ and $q(\bm L')$ are distinct.\label{def:contextual_mapping_rule2}
   \end{enumerate}
\end{definition}
Intuitively, a contextual map can be thought of as a function that outputs unique ``id-values''. The only way for a token (column) in \( \bm {L} \subset \mathbb{R}^{d\times n} \) to share an ``id-value'' (element of \( q(\bm {L}) \)) is to map the exact same sequence. As each token in the sequence is mapped to a unique value, an appropriately constructed feed-forward neural network can map a sequence to any other desired sequence, providing a universal approximation guarantee.
In \citet{yun2020n}, such a contextual map is implemented via \emph{selective shift operators} and \emph{all-max-shift operators} through careful construction of multi-head self-attention layers.



\section{Methodology}


We propose a variation of the sparse attention transformer -- sampled sparse attention transformer -- applicable to point sets. We deviate from the typical sparse attention transformer in two ways. First, we randomly sub-sample the input point set \( l \) times, with each sub-sample being evaluated through a shared multi-head self-attention layer. Secondly, we propose a simple Hamiltonian self-attention mechanism, a special case of the sparse attention mechanism, to reduce the computation complexity of considering point sets. This ultimately yields the variant of the typical sparse transformer (Eq.~\ref{eq:dense_tf}) which can be interpreted as using a \emph{sampled attention} mechanism, as depicted in Fig.~\ref{fig:diagram}. To study the approximation capabilities of our proposed architecture, we prove that our sampled sparse attention transformer is a universal approximator of set-to-set functions.

\def\SUBSETNOTATION{1}

\ifnum\SUBSETNOTATION=1
\subsection{Random Element Sampling}
For a point set input \( \bm X \in \R^{d \times n} \), instead of directly applying the transformer attention layer to $n$ tokens, we process \( l \) many sub-sampled inputs \( \subX^{i} \in \R^{d \times n_{s}} \) for \( i \in [l] \) and \( 2 \leq n_{s} \leq n \).
For simplicity, we assume that \( (n_{s} - 1) \cdot l = n \).
The sub-sampled inputs \( \subX^{i} \) can be defined by taking various column submatrices:
%
\begin{align}\label{eq:random_element_sampling}
    &\subX^{i} = \bm X_{\mathcal{R}^{i} \cup \mathcal{R}_{1}^{\gamma(i)}}; \quad \gamma(v) = 1 + (v \mod l),
%
\end{align}
where \( \mathcal{R}^{1}, \ldots \mathcal{R}^{l} \) are randomly selected ordered index sets, such that \( \vert \mathcal{R}^{i} \vert = n_{s} - 1 \) and \( \mathcal{R}^{i} \cap \mathcal{R}^{j} = \emptyset \) for \( i \neq j \). The index element \( \mathcal{R}_{1}^{i} \) denotes the first index in the ordered set \( \mathcal{R}^{i} \). The \emph{cycle} function \( \gamma: [l] \rightarrow [l] \) ensures that the edge-case of \( \subX^{l} \) is well defined, \emph{i.e.}, \( \gamma(l) = 1 \).

Intuitively, the sequence of sub-sampled inputs \( \subX^{1}, \ldots, \subX^{l} \) can be interpreted as a rolling window of \( (n_{s} - 1) \cdot l = n \) many sampled point set elements. Indeed, by concatenating the index sets in order, \( \subX^{i} \) is a sliding window of the elements with size \( n_{s} \) and stride \( n_{s} - 1 \) (with wrapping).

It should be noted that \( \subX^{i} \) can be treated as a random variable. As such a singular realization of the sampled elements can be viewed as a Monte Carlo sample over the set of ordered point sequences~\citep{metropolis1949monte}.
Computationally, by applying a dense self-attention layer to each of the sub-sampled elements \( \subX^{i} \), the total complexity of evaluating \( l \) many self-attention layer is \( O(l \cdot n_{s}^{2}) \). We however note that the \( l \) self-attention layers can be evaluated in parallel, which yields a trade-off between individual self-attention complexity \( O(n_s^{2}) \) and computation time.

To gain intuition, consider the ``limiting behaviours'' of our random element sampling: taking \( n_{s} = n + 1 \) can be interpreted as taking the whole sequence with \( l = 1 \), \emph{i.e.}, \( \subX^{1} = \bm X \) which under dense attention would result in complexity \( O(n^2) \).
On the other end, if we take \( n_{s} = 2 \), we get \( l = n \) pairs of points \( \vert \subX^i \vert = 2 \); processing every such pair with dense self-attention results in \( n \) many \( O(1) \) self-attention evaluations. Random element sampling with dense attention layers can be interpreted as an instance of sparse attention, see Fig.~\ref{fig:after_sampling}. 

\else

\subsection{Random Element Sampling}
For a point set input \( \bm X \in \R^{d \times n} \), instead of directly inputting $n$ tokens into the transformer's attention layer, we process \( l \) many sub-sampled inputs \( \subX^{i} \in \R^{d \times n_{s}} \) for \( i \in [l] \). We assume that \( n_{s} l = n \) for simplicity and that \( n_{s} \ll n \).
Specifically, we define \( \subX^{i} \) as the following:
%
\begin{align}
    &\subX^1 = \bm X \bm \U^{1}_{n_s}; \quad \subX^l = \subX^{l-1} \bmshiftl + \bm X [\zeros_n^T, \bm \U_{n_s-2}^l, \zeros_n^T] +  \subX^{1} \bmshiftu \nonumber\\
    &\subX^i = \subX^{i-1} \bmshiftl + \bm X \bm [\zeros_n^T, \bm \U_{n_s-1}^i], \quad \text{ for } i \in \{2, 3,\dotso, l-1\},
\end{align}
where \( \bm \U_{a}^{.} \in \R^{n \times a} \) such that each row of \( [(\bm \U^{1}_{n_s})^T, (\bm \U^{2}_{n_s-1})^{T}, \ldots, (\bm \U^{l-1}_{n_s-1})^{T}, (\bm \U^{l}_{n_s-2})^{T}] \) consists of one-hot vectors generated by sampling without replacement from the set of all \( n \)-length one-hot vectors. The upper-shift (lower-shift) matrix \( \bmshiftu \) (\(\bmshiftl\)) is defined as the matrix in \( \R^{n \times n} \) which consists of all zeros except for its upper-right (bottom-left) most element \( \shiftu_{1,n_{s}} = 1 \) (\( \shiftl_{n_{s},1} = 1 \)).

Intuitively, the shift operators ``extracts'' a column when applied to a matrix. For instance given \( \bm M \in \R^{n_{s} \times n_{s}} \), we can simply \( {\bm M} \bmshiftl = [{\bm M}_{n_{s}}, \zeros_{n_{s}}^{T}, \ldots, \zeros_{n_{s}}^{T}] \) and \( {\bm M} \bmshiftu = [\zeros_{n_{s}}^{T}, \ldots, \zeros_{n_{s}}^{T}, {\bm M}_{1}] \) -- \emph{i.e.}, the lower-shift operator extracts the last column of elements and the upper-shift operator extracts the first column.
Furthermore, the sequence of sub-sampled inputs \( \subX^{i} \) can be interpreted as a rolling window of sampled point set elements. Indeed, the input \( \bm X \) matrix multiplied to a matrix whose columns consists of one-hot vectors simply extracts elements, \emph{i.e.}, \( {\bm X} (\indicator_{n}^{k})^T = {\bm X}_{k} \),
Thus, for \( i \in \{2, 3, \ldots, l-1 \} \) the sub-sample \( \subX^{i} \) consists of \( n_{s} - 1 \) new (w.r.t. point seen in the sequence) elements sampled from (columns of) \( \bm X \) and one old element \( \subX^{i-1}_{n_{s}} \). The final element of the sequence also has an element from the previous sample in the sequence \( \subX^{l-1}_{n_{s}} \), but also first element of the first sampled points \( \subX^{1}_{n_{s}} \).

\todo{Commented out below was discussion regarding a position embedding. I am unsure what this is referred to. I think we can notate this in the dense transformer definition regardless.}

\todo{Maybe consider moving this in the section where the entire transformer is introduced}

The proposed sampling of elements of \( \bm X \) can be viewed as a Monte Carlo simulation \citep{metropolis1949monte} over the set of ordered point sequences (of length \( n_{s} + (l-2)(n_{s} - 1) + (n_{s} - 2) = (l+1)(n_{s} - 1) \)).
\todo{Practically, sequences are sampled for different batches and epochs, and with increasing number of the deterministic orders are learned and aggregated by the transformer, the untouchable element-wise relations within a set are learned.}
Computationally, by applying a dense self-attention layer to the sub-sampled elements \( \subX^{i} \), we have a trade-off between the number of subsets \( l \) and the complexity of evaluating the self-attention layers \( O(n_s^{2}) \) (by assumption we have \( l \ll n_{s} \)).
When $l=1$ we have \( n_{s} = n \), the whole point sequence is passed to the self-attention layer, resulting in complexity $O(n^2)$, which is just the original input \( \bm X \) applied to a dense transformer.
When $l=n$ we have \( n_{s} = 1 \), $n$ \todo{points-pairs are passed to the self-attention layer with complexity $O(4)$, so the self-attention is trained $n$ times.}

\fi

\subsection{Hamiltonian Self-Attention}

The random element sampling discussed in the previous section reduces the computational complexity of dense self-attention-layers from \( O(n^2) \) to \( O(l \cdot n_{s}^2) = O(n^2 / l) \) (as \( (n_{s} - 1) \cdot l = n \)) by processing each sampled set of points \( \subX^{i} \) through individual self-attention layers. Despite this improved computational complexity, the quadratic scaling of \( n \) can still be costly for point clouds.


As such, instead of evaluating each sampled element \( \subX^{1}, \ldots, \subX^{l} \) with a dense self-attention layer, we propose a sparse attention layer. 
Sparse attention mechanisms can be formally defined via the attention patterns \( \{ \mathcal{A}_{k} \}_{k \in [n_{s}]} \), where \( j \in \mathcal{A}_{k} \) implies that the \( j \)-th token will attend to the \( k \)-th token.
We propose the use of an attention mechanism, dubbed as \emph{Hamiltonian self-attention}, which is defined by the following attention patterns:
%
%
\begin{align}\label{eq:attention_pattern}
   \mathcal{A}_k = \begin{cases}
        \{k, k + 1\} & \textrm{if } 1 \leq k < n_s \\
        \{ k\} & \textrm{otherwise } k=n_s
   \end{cases},
\end{align}
which ensures that the set of attention patterns \( \{ \mathcal{A}_{k} \}_{k \in [n_{s}]} \) define a \emph{Hamiltonian path}.
%
Indeed, if we fix a subset of elements \( \subX^{i} \), by starting at \( \subX^{i}_{1} \) and following the attended elements (ignoring self-attention \( k \in \mathcal{A}_{k} \)), we visit every token exactly once. Fig.~\ref{fig:Hamiltonian} shows the corresponding attention matrix, where the Hamiltonian path corresponds to off-diagonal elements and self-attention corresponds  to the diagonal elements, respectively.

For Hamiltonian self-attention, computing the attention mechanism according to Eq.~\ref{eq:attention_pattern} only requires \( 2n_{s} = O(n_{s}) \) many evaluations. Thus by using our proposed sparse attention for each \( \subX^{1}, \ldots, \subX^{l} \), in comparison to dense attention, the computational complexity reduces from \( O(n^{2} / l^{2}) \) to \( O(n / l) \).


The proposed Hamiltonian self-attention mechanism is rather simple and general. For instance, in the general case sparsity patterns can be defined for each individual layer (resulting in an addition superscript for each \( A_{k} \)).
Despite this, the attention patterns \( \{ A_{k} \}_{k \in [n_{s}]} \) satisfy important key assumptions for proving that the attention pattern will result in a sparse transformer that is a universal approximator~\citep[Assumption 1]{yun2020n}. In particular, by stacking \( (n_{s} - 1) \) many attention layers, our Hamiltonian self-attention will allow any element to indirectly or directly attend all other element in a \( \subX^{i} \).
The proposed Hamiltonian self-attention could also be viewed as a special case of window attention in \citet{zaheer2020big}, where elements are linked undirectedly.


\subsection{Sampled Sparse Attention Transformer}\label{sec:sampled_attention}

Given the setup of random element sampling and Hamiltonian self-attention, we can define our proposed sampled transformer for continuous set-to-set function approximation:
%
\begingroup
\allowdisplaybreaks
\begin{subequations}
    \begin{align}
       &\shead_k^{j}(\subX^i) = (\bm W^j_V \subX^{i}_{\mathcal{A}_k}) \cdot \sigma_S [ (\bm W^j_K \subX^{i}_{\mathcal{A}_k})^T \bm W^j_Q \subX^i_k]\\\label{eq:g_func}
       &g^i (\subX^i) = \subX^i + \bm W_O \begin{bmatrix} \shead^{1}(\subX^i) \\ \vdots \\ \shead^{h}(\subX^i) \end{bmatrix}\\\label{eq:sampled_attn}
       &\sattn(\bm X) = g^{l}(\subX^l) \circ g^{l-1}(\subX^{l-1}) \circ \cdots \circ g^1(\subX^1)\\\label{seq:stacked_sampled_attn}
       &\stb(\bm X) = \sattn(\bm X) + \bm W_2 \cdot \text{ReLU}\left(\bm W_1 \sattn(\bm X)\right).
    \end{align}%
    \label{eq:sampled_tf}%
 \end{subequations}%
 \endgroup
In Eq.~\ref{eq:sampled_attn}, composition is w.r.t. the induced linear maps from matrices given by Eq.~\ref{eq:g_func}.
The learnable parameters of the sampled transformer are the same as the usual dense transformer in Eq.~\ref{eq:dense_tf}.

As the attention pattern of each \( \subX^{i} \) forms a Hamiltonian path, and each \( \subX^{i} \) shares an element with the proceeding \( \subX^{\gamma(i)} \), the joint attention map makes a \textit{Hamiltonian cycle} path. In other words, the shared index \( \mathcal{R}_{1}^{\gamma(i+1)} \) in Eq.~\ref{eq:random_element_sampling} links each individual Hamiltonian path given by Eq.~\ref{eq:attention_pattern}, leading the attention matrix to form a \textit{cycle attention} as shown in Fig.~\ref{fig:cycle}. 
Furthermore, the permutation of elements in cycle attention corresponds to the swapping of nodes in the Hamiltonian cycle, with corresponding links and swapping of element values in the attention matrix, see in Fig.~\ref{fig:swap}.
As a result, the combined randomization from using random element sampling and Hamiltonian self-attention can be thought of as sampling from the set of Hamiltonian cycle graphs from the complete attention graph, resulting in the \textit{sampled attention} depicted in Fig.~\ref{fig:sampled}. 

Unlike dense attention, sparse attention patterns are not generally permutation invariant. Indeed, if we permute the columns of \( \subX^{i} \), the elements attended according to \( \{ \mathcal{A} \}_{k \in [n_{s}]} \) are not the same. As such, applying \( \{ \mathcal{A}_{k} \}_{k \in [n_{s}]} \) directly to \( \bm X \) is not valid for point clouds, which requires a permutation invariant operation. However, in our case the sparse attention heads are being applied to \emph{randomized} sub-sampled element sets \( \subX^{i} \). Ignoring computation, if we continue to sample the randomized elements \( \subX^{i} \) and average the resulting attention (w.r.t. the entire point set \( \bm X \)), the attention will converge to dense attention -- through randomization of \( \subX^{i} \), the event that any non-self-edge appears in a sampled attention graph (as per Eq.~\ref{eq:attention_pattern}) is equiprobable.
This also holds when fixing the order of elements while applying randomly sampled Hamiltonian cycle attention.
As such, the sampled transformer can be used to approximate a permutation invariant operator, and thus be used to approximate set-to-set functions.

Of course, sampling sufficiently many realizations of Hamiltonian cycle attention to converge to dense attention is impractical. Instead, in practice, we re-sample the attention pattern only for each batch and epoch. 
Although this may seem like a crude approximation to dense attention, similar methods are successful in Dropout~\citep{srivastava2014dropout}, which even induces desirable model regularization. Furthermore, our empirical results indicate that sampled sparse attention closely approximates the more expensive (and infeasible at the typical point set scales) dense attention.



\subsection{Sampled Transformer as a Universal Approximator}
We formally guarantee the representation power of the proposed sampled transformer by proving universal approximation for set-to-set functions.
As our sampled transformer Eq.~\ref{eq:sampled_attn} is similar to dense / sparse transformers presented by \citet{yun2019transformers,yun2020n}, we follow their framework (Sec.~\ref{subsec:ua}) to prove our universal approximation property.

\begin{corollary}[Sampled Transformer is a Universal Approximator]\label{thm:our_ua}
    There exist sampled (sparse) Transformers that are universal approximators in the sense of Theorem~\ref{theorem:ua}.
\end{corollary}%
%
To prove our Corollary, we extend the proof of \citet{yun2019transformers,yun2020n} by showing that our sparse attention mechanisms with random element sampling can also implement a selective shift operator. As a result, we show that the proposed sampled sparse attention transformer is a universal approximator in the context of set-to-set functions.
See \S\ref{app:ua} in the supplementary material for the full proof of the universal approximation property.

\begin{table}[t]
    \caption{Object classification on ModelNet40. Here [ST] denotes that model adopts the standard (dense) transformer, while [T] denotes all other transformers.}
    \label{tab:classification}
    \begin{subtable}[h]{0.5\textwidth}
        \centering
        \resizebox{0.85\textwidth}{!}{%
        \begin{tabular}{l|c}
            \toprule
            Supervised Methods & Accuarcy \\
            \midrule
            PointNet~\citep{qi2017pointnet} & 89.2\% \\
            PointNet++~\citep{qi2017pointnet++} & 90.7\% \\
            PointCNN~\citep{li2018pointcnn} & 92.5\% \\
            KPConv~\citep{thomas2019kpconv} & 92.9\% \\
            DGCNN~\citep{wang2021unsupervised} & 92.9\% \\
            RS-CNN~\citep{liu2019relation} & 92.9\% \\
            \/[T\/] PCT~\citep{guo2021pct} & 93.2\% \\
            \/[T\/] PVT~\citep{zhang2021pvt} & 93.6\% \\
            \/[T\/] PointTransformer~\citep{zhao2021point} & 93.7\% \\
            \/[T\/] Transformer~\citep{yu2022point} & 91.4\% \\
            \bottomrule
        \end{tabular}
        }
    \end{subtable}
    \hfill
    \begin{subtable}[h]{0.5\textwidth}
        \centering
        \resizebox{0.85\textwidth}{!}{%
        \begin{tabular}{l|c}
            \toprule
            Self-Supervised Methods & Accuarcy \\
            \midrule
            OcCo~\citep{wang2021unsupervised} & 93.0\% \\
            STRL~\citep{huang2021spatio} & 93.1\% \\
            IAE~\citep{yan2022implicit} & 93.7\% \\
            \/[ST\/]Transformer-OcCo~\citep{yu2022point} & 92.1\% \\
            \/[ST\/]Point-BERT~\citep{yu2022point} & 93.2\% \\
            \/[ST\/]Point-MAE~\citep{pang2022masked} & 93.8\% \\
            \midrule
            \/[ST\/]\textbf{MAE-dense (ours)} & 93.6\%\\
            \/[T\/]\textbf{MAE-sampled (ours)} & 93.7\%\\
            \bottomrule
        \end{tabular}
        }
    \end{subtable}
\end{table}
\section{Experiments}\label{sec:exp}
We evaluate our proposed sampled attention in popular transformer-based frameworks as well as basic settings.
To begin with, we compare our sampled attention (Fig.~\ref{fig:sampled}) with dense attention via the pre-training and fine-tuning framework~\citep{yu2022point,pang2022masked}, where we pre-train our model on ShapeNet~\citep{chang2015shapenet} via the reconstruction task, and further evaluate the performance on three downstream fine-tuning tasks: classification, transfer learning, and few-shot learning in ModelNet40~\citep{wu20153d} or ScanObjectNN~\citep{uy2019revisiting}.
In addition, to eliminate the influence of other factors, we compared the dense, sparse, sampled, and $k$NN attention (Definition~\ref{def:knn_atten}), together with other sparse transformer such as Inducting Points \citep{lee2019set} and Stratified Strategy \citep{lai2022stratified}, in a basic classification setting consisting of a transformer block with a single attention layer for feature aggregation.
Further, we compare the sampled attention with the $k$NN attention in the hierarchical grouping and merging structure following the Point-Transformer~\citep{zhao2021point}.
Finally, we test the proposed sampled attention in 2D set datasets introduced by \citet{lee2019set}.

\begin{table}[t]
    \caption{%
        Transfer learning on the classification task, measured by the Accuracy (\%).
    }%
    \label{tab:transfer_learning}
    \centering
        \resizebox{0.5\textwidth}{!}{%
        \begin{tabular}{l|ccc}
            \toprule
            Methods & OBJ-BG & OBJ-ONLY & PB-T50-RS \\
            \midrule
            PointNet~\citep{qi2017pointnet} & 73.3 & 79.2 & 68.0 \\
            SpiderCNN~\citep{xu2018spidercnn} & 77.1 & 79.5 & 73.7 \\
            PointNet++~\citep{qi2017pointnet++} & 82.3 & 84.3 & 77.9 \\
            PointCNN~\citep{li2018pointcnn} & 86.1 & 85.5 & 78.5 \\
            DGCNN\citep{wang2021unsupervised} & 82.8 & 86.2 & 78.1 \\
            BGA-DGCNN~\citep{uy2019revisiting} & - & - & 79.7 \\
            BGA-PN++~\citep{uy2019revisiting} & - & - & 80.2\\
            Point-BERT~\citep{yu2022point} & 87.43 & 88.12 & 83.07 \\
            Point-MAE~\citep{pang2022masked} & 90.02 & 88.29 & \textbf{85.18} \\
            \midrule
            \textbf{MAE-dense (ours)} & \textbf{90.36} & 88.50 & 83.41 \\
            \textbf{MAE-sampled (ours)} & 89.68 & \textbf{88.81} & 82.44 \\
            \bottomrule
        \end{tabular}
    }
\end{table}
\subsection{Comparsion on Pre-training and Fine-tuning Framework}\label{subsec:bert}
\paragraph{Pre-training.}
We adopted the masked auto-encoder (MAE)~\citep{he2022masked} to process the point cloud data (denoted as MAE-dense) for pre-training, which is close with Point-MAE~\citep{pang2022masked}. Note that MAE-dense adopts dense-attention layers in its encoder and decoder network. To evaluate the effectiveness of our claimed contribution, we replace the dense-attention layer in MAE-dense with our sampled-attention layer (Fig.~\ref{fig:sampled}) while keeping the other components fixed. It is denoted as MAE-sampled. 

\begin{table*}[t]
    \caption{%
        Mean $\pm$ std. dev. accuracy (\%) for 10 independent Few-shot classification experiments.
    }%
    \label{tab:few_shot}
    \centering
        \resizebox{0.95\textwidth}{!}{%
        \begin{tabular}{l|cccc}
            \toprule
            Methods & 5-way, 10-shot & 5-way,20-shot & 10-way,10-shot & 10-way, 20-shot \\
            \midrule
            DGCNN-rand~\citep{wang2021unsupervised} & 31.6 $\pm$ 2.8 & 40.8 $\pm$ 4.6 & 19.9 $\pm$ 2.1 & 16.9 $\pm$ 1.5 \\
            DGCNN-OcCo~\citep{wang2021unsupervised} & 90.6 $\pm$ 2.8 & 92.5 $\pm$ 1.9 & 82.9 $\pm$ 1.3 & 86.5 $\pm$ 2.2 \\
            Transformer-rand~\citep{yu2022point} & 87.8 $\pm$ 5.2 & 93.3 $\pm$ 4.3 & 84.6 $\pm$ 5.5 & 89.4 $\pm$ 6.3 \\
            Transformer-OcCo~\citep{yu2022point} & 94.0 $\pm$ 3.6 & 95.9 $\pm$ 2.3 & 89.4 $\pm$ 5.1 & 92.4 $\pm$ 4.6 \\
            Point-BERT~\citep{yu2022point} & 94.6 $\pm$ 3.1 & 96.3 $\pm$ 2.7 & 91.0 $\pm$ 5.4 & 92.7 $\pm$ 5.1 \\
            Point-MAE~\citep{pang2022masked} & 96.3 $\pm$ 2.5 & 97.8 $\pm$ 1.8 & 92.6 $\pm$ 4.1 & \textbf{95.0 $\pm$ 3.0} \\
            \midrule
            \textbf{MAE-dense (ours)} & 95.9 $\pm$ 3.1 & 97.2 $\pm$ 2.1 & 90.8 $\pm$ 5.0 & 92.8 $\pm$ 3.9\\
            \textbf{MAE-sampled (ours)} & \textbf{97.0 $\pm$ 2.3} & \textbf{98.3 $\pm$ 1.6} & \textbf{92.7 $\pm$ 5.4} & 93.8 $\pm$ 3.5\\
            \bottomrule
        \end{tabular}
    }
\end{table*}
To pre-train the MAE-dense and MAE-sampled, we first follow the standard train-test split of ShapeNet~\citep{chang2015shapenet} adopted by \citet{pang2022masked,yu2022point}.
Further, the Furthest Points Sampling (FPS) and nearest neighbour search were adopted in tokenization~\citep{yu2022point} step, which means each input point cloud consisting of 1024 points was divided into 64 groups / tokens of size 32 points each.
Tokens were further mapped to 256-dimensional latent vectors by MLP layers and max-pooling.
In addition, we have 12 stacked transformers in the encoder (masking ratio of 70\%) and 1 single transformer in the decoder, both with $h=8$, $d=32$ and $r=256$.
The batch size is 64 and the epoch number is 300. 
We used the AdamW~\citep{loshchilov2017decoupled} optimizer with cosine learning rate decay~\citep{loshchilov2016sgdr}, an initial learning rate of 0.0005, and weight decay of 0.05.

\paragraph{Classification}
The pre-trained MAE-dense and MAE-sampled models are first evaluated on the classification task in ModelNet40 \citep{wu20153d}.
Specifically, we build the classifier by keeping the encoder structure and weights of the pre-trained MAE-dense and MAE-sampled models, followed by max-pooling as well as a fully connected layer of dimension $[256, 256, 40]$ to map the global token of a dimension of 256 to the 40 categories. Similar to ~\citet{yu2022point}, we further data-augment the point cloud training set via random scaling and translation during training.
As shown in Tab.~\ref{tab:classification}, the proposed method achieved the second best performance compared with the most recent state-of-the-arts.
Our sampled attention can achieve 
an accuracy improvement of \( 0.1\% \) when
compared to dense attention, while reducing the complexity from $O(n^2)$ to $O(n)$.



\paragraph{Transfer Learning}
We additionally included the transfer learning as a fine-tuning classification task, which is implemented on the ScanObjectNN~\citep{uy2019revisiting} dataset with 2902 point clouds from 15 categories.
We follow the data pre-processing and fine-tuning setting from Point-BERT~\citep{yu2022point} with the same three variants: OBJ-BG, OBJ-ONLY, and PB-T50-RS.
As we can see in Tab.~\ref{tab:transfer_learning}, our sampled attention achieved a competitive performance in comparison with dense attention while reaching state-of-the-art performance.

\paragraph{Few Shot Learning}
The pre-trained MAE-dense and MAE-sampled models are finally evaluated on a few shot learning task. Following \citet{sharma2020self,wang2021unsupervised,yu2022point,pang2022masked}, the few-shot learning adopted an $k$-way, $m$-shot training setting on the ModelNet40~\citep{wu20153d} dataset, where $k$ represents the number of randomly sampled classes and $m$ the number of randomly sampled examples per class.
The testing split is 20 randomly sampled unseen examples from each class.
We set $k\in\{5, 10\}$ and $m\in\{10,20\}$, and report the mean accuracy with standard deviation for 10 independent experiments.
As shown in Tab.~\ref{tab:few_shot}, our proposed MAE-sampled outperformed all state-of-the-art methods on 3 out of 4 settings, while MAE-sampled consistently outperformed MAE-dense.

\subsection{Comparsion on Basic Classification Setting}\label{subsec:basic_setting}
\begin{table*}[t]
    \caption{Object classification accuracy (\%) for different attention mechanisms in the basic setting.
    OM denotes \textit{out of memory}.}%
    \label{tab:single_layer}%
    \centering
        \resizebox{0.95\textwidth}{!}{%
        \begin{tabular}{l|cccccccc}
            \toprule
            \#Points & 256 & 512 & 768 & 1024 & 2048 & 3072 & 4096 & 8192 \\
            \midrule
            MLP + FC (no attention) & 85.96 & 86.24 & 85.43 & 85.96 & 85.84 & 86.61 & 86.32  & 86.13 \\
            Dense Attention & 87.78 & 88.72 & 88.11 & 88.47 & 88.39 & OM & OM  & OM  \\
            \midrule
            Inducting Points \citep{lee2019set} & 84.21 & 81.25 & 82.55 & 81.57 & 80.96 & 76.18 & 75.13 & 75.65 \\
            Stratified Strategy \citep{lai2022stratified} & 87.21 & 87.62 & 86.69 & 85.99 & 85.34 & 84.32 & OM & OM \\
            Sparse Attention & 87.09 & \textbf{88.03} & 87.54 & 87.74 & 87.58 & 87.42 & 87.42 & 87.58 \\
            $k$NN Attention & 85.80 & 84.74 & 85.35 & 84.70 & 82.95 & 82.58 & 82.26 & OM \\
            Sampled Attention &\textbf{87.34} & 87.93 & \textbf{87.66} & \textbf{88.03} & \textbf{87.82} & \textbf{87.18} & \textbf{87.46} & \textbf{87.73} \\
            \bottomrule
        \end{tabular}
        }
\end{table*}
Our inputs are clouds of $n$ points with 3D coordinates as position and its normal information as features.
The feature and position are first transformed by two separate MLP layers with hidden dimensions $[64, 256]$, and then added together as the input of a single layer transformer with $h=8$, $r=256$, and $d=32$, as per Eq.~\ref{eq:dense_tf}~and Eq.~\ref{eq:sampled_tf}.
The transformer output of $\mathbb{R}^{n\times 256}$ is then summarized by max-pooling to obtain a global feature with a dimension of 256, followed by a fully connected layer to map it to the category vector.
Here we tested this basic pipeline with $n \in \{256, 512, 768, 1024, 2048, 3072, 4096, 8192\}$ for each of the dense, sparse, $k$NN, the proposed sampled attention layers, Inducting Points \citep{lee2019set}, and Stratified \citep{lai2022stratified}, including an additional case without attention layer (MLP+Full Connected layer) as the baseline.

We addiitonally included the emeory usage for different attention layers in \S~\ref{app:memory_usage} Tab.~\ref{tab:memory}.
As shown in Tab.~\ref{tab:single_layer} and Tab.~\ref{tab:memory}, the model with dense attention layers achieves the best performance as it considers all $O(n^2)$ connections directly with relatively few parameters to train. 
However, it runs out of the 24 Gigabytes memory when the number of points $n \geq 3072$, due to the quadratic complexity.
While both sparse and sampled transformers have a computational complexity of $O(n)$, our model with sampled attention outperformed the sparse one, in line with the strong theoretical guarantees we provide. 
We conjecture that the improvements of sampled transformer over the sparse transformer may indicate that the additional randomness (randomly shuffling points, w / o attention) leads to a better approximation of the $O(n^2)$ connections in a manner analogous to Dropout~\citep{srivastava2014dropout}. 
In addition, the transformer with $k$NN attention layers has the worst performance, as the permutation could not extend its receptive field.
Finally, the proposed sampled attention layer also outperforms existing point-cloud-oriented sparse attentions, such as Inducting Points \citep{lee2019set}, and Stratified \citep{lai2022stratified}.
Details of the comparsion could be found in \S~\ref{app:inducting} and \S~\ref{app:stratified}, respectively.

\subsection{Comparsion on Hierarchical Transformer Structure}
\begin{table}[t]
    \caption{%
        Classification accuracy (\%) for sampled and $k$NN attention with hierarchical model structure.%
    }%
    \label{tab:hier_tf}%
    \centering
        \resizebox{0.45\textwidth}{!}{%
        \begin{tabular}{l|ccccc}
            \toprule
            \#Layers & 1 & 2 & 3 & 4 & 5 \\
            \midrule
            sampled attention & \textbf{74.55} & \textbf{88.0} & \textbf{90.5} & \textbf{91.0} & \textbf{91.8} \\
            $k$NN attention & 66.23 & 82.8 & 90.1 & \textbf{91.0} & 91.4 \\
            \bottomrule
        \end{tabular}
        }
\end{table}
We further compare our sampled attention with $k$NN attention by adopting the hierarchical structure for the classification task under the framework of \cite{zhao2021point}.
Each hierarchical layer is obtained by FPS,
followed by the nearest neighbour search for the grouping, using MLPs with max-pooling for feature merging, and transformers for feature mapping.
The grouping stage within each hierarchical layer summarizes the point cloud into key (subset) points.

The total hierarchical layer number is $t=5$, the parameters for which we chose the number $k$ of nearest neighbours \{8, 16, 16, 16, 16\}, strides \{4, 4, 4, 4, 4\}, self-attention feature dimensions \{32, 64, 128, 256, 512\}, and transformer blocks \{2, 3, 4, 6, 3\}. The scalar attention (Eq.~\ref{eq:dense_tf} or Eq.~\ref{eq:sampled_tf}) is adopted specifically for comparison.
Results shown in Tab.~\ref{tab:hier_tf} demonstrate that our sampled attention outperforms the $k$NN attention in line with our randomly sampled receptive field.
Furthermore, the performance of the $k$NN layer improved greatly from $t=1$ to $t=2$ and from $t=2$ to $t=3$ as its receptive field extends due to the multiple hierarchical layers.
Finally, $k$NN with vector attention~\citep{yu2022point} (reported in Tab.~\ref{tab:classification} on the PointTransformer row) achieved a better performance, in line with the observation that replacing the softmax with learnable MLPs $\gamma$ in the transformer can easier make $k$NN attention a universal approximator of continuous functions.
Detailed analysis is provided in \S\ref{app:ua_spth} in the supplementary material.
The performance difference between scalar attention and vector attention is shown in the Tab. 7 of \cite{yu2022point}, and is also analyzed in \cite{yun2020n}.


\section{Amortized Clustering}
We test the proposed sampled attention in 2D set datasets in the encoding-decoding framework introduced by \citet{lee2019set}. And the task is about 
using a neural network to learn the parameters of the mixture Gaussian distribution from the input set data.
As we can see in Tab.~\ref{tab:clustering}, the sampled attention could be a plug-in module to replace the dense attention in the inducting points structure with competitive performance but theoretically less computational complexity.
Detailed implementation and comparsion could be found in \S~\ref{app:clustering}.

\section{Conclusion}
In this paper, we present an $O(n)$ complexity sparse transformer -- \textit{sampled transformer} -- which directly handles point set data.
By relating the permutation of set elements to the sampling of Hamiltonian cycle attention, we relieve the model of inappropriate permutation variance.
The result is a sampled attention scheme that implements Monte Carlo simulation to approximate a dense attention layer with a prohibitive $O(n^2)$ number of connections.
To guarantee the representation power of the proposed sampled transformer, we showed that it is a universal approximator of set-to-set functions. 
Motivated also by the strong empirical performance that our model achieves, we hope this work will help to shed light on the sparse transformer in dealing with sets.

\bibliography{example_paper}

\begin{thebibliography}{51}
\providecommand{\natexlab}[1]{#1}
\providecommand{\url}[1]{\texttt{#1}}
\expandafter\ifx\csname urlstyle\endcsname\relax
  \providecommand{\doi}[1]{doi: #1}\else
  \providecommand{\doi}{doi: \begingroup \urlstyle{rm}\Url}\fi

\bibitem[Bao et~al.(2021)Bao, Dong, and Wei]{bao2021beit}
Bao, H., Dong, L., and Wei, F.
\newblock Beit: Bert pre-training of image transformers.
\newblock \emph{arXiv preprint arXiv:2106.08254}, 2021.

\bibitem[Buitinck et~al.(2013)Buitinck, Louppe, Blondel, Pedregosa, Mueller,
  Grisel, Niculae, Prettenhofer, Gramfort, Grobler, Layton, VanderPlas, Joly,
  Holt, and Varoquaux]{sklearn_api}
Buitinck, L., Louppe, G., Blondel, M., Pedregosa, F., Mueller, A., Grisel, O.,
  Niculae, V., Prettenhofer, P., Gramfort, A., Grobler, J., Layton, R.,
  VanderPlas, J., Joly, A., Holt, B., and Varoquaux, G.
\newblock {API} design for machine learning software: experiences from the
  scikit-learn project.
\newblock In \emph{ECML PKDD Workshop: Languages for Data Mining and Machine
  Learning}, pp.\  108--122, 2013.

\bibitem[Chang et~al.(2015)Chang, Funkhouser, Guibas, Hanrahan, Huang, Li,
  Savarese, Savva, Song, Su, et~al.]{chang2015shapenet}
Chang, A.~X., Funkhouser, T., Guibas, L., Hanrahan, P., Huang, Q., Li, Z.,
  Savarese, S., Savva, M., Song, S., Su, H., et~al.
\newblock Shapenet: An information-rich 3d model repository.
\newblock \emph{arXiv preprint arXiv:1512.03012}, 2015.

\bibitem[Dai et~al.(2019)Dai, Yang, Yang, Carbonell, Le, and
  Salakhutdinov]{dai2019transformer}
Dai, Z., Yang, Z., Yang, Y., Carbonell, J., Le, Q.~V., and Salakhutdinov, R.
\newblock Transformer-xl: Attentive language models beyond a fixed-length
  context.
\newblock \emph{arXiv preprint arXiv:1901.02860}, 2019.

\bibitem[Dosovitskiy et~al.(2020)Dosovitskiy, Beyer, Kolesnikov, Weissenborn,
  Zhai, Unterthiner, Dehghani, Minderer, Heigold, Gelly,
  et~al.]{dosovitskiy2020image}
Dosovitskiy, A., Beyer, L., Kolesnikov, A., Weissenborn, D., Zhai, X.,
  Unterthiner, T., Dehghani, M., Minderer, M., Heigold, G., Gelly, S., et~al.
\newblock An image is worth 16x16 words: Transformers for image recognition at
  scale.
\newblock \emph{arXiv preprint arXiv:2010.11929}, 2020.

\bibitem[Fuchs et~al.(2020)Fuchs, Worrall, Fischer, and Welling]{fuchs2020se}
Fuchs, F., Worrall, D., Fischer, V., and Welling, M.
\newblock Se (3)-transformers: 3d roto-translation equivariant attention
  networks.
\newblock \emph{Advances in Neural Information Processing Systems},
  33:\penalty0 1970--1981, 2020.

\bibitem[Guo et~al.(2021)Guo, Cai, Liu, Mu, Martin, and Hu]{guo2021pct}
Guo, M.-H., Cai, J.-X., Liu, Z.-N., Mu, T.-J., Martin, R.~R., and Hu, S.-M.
\newblock Pct: Point cloud transformer.
\newblock \emph{Computational Visual Media}, 7\penalty0 (2):\penalty0 187--199,
  2021.

\bibitem[Guo et~al.(2019)Guo, Qiu, Liu, Shao, Xue, and Zhang]{guo2019star}
Guo, Q., Qiu, X., Liu, P., Shao, Y., Xue, X., and Zhang, Z.
\newblock Star-transformer.
\newblock \emph{arXiv preprint arXiv:1902.09113}, 2019.

\bibitem[Han et~al.(2022)Han, Jin, Cheng, and Xiao]{han2022dual}
Han, X.-F., Jin, Y.-F., Cheng, H.-X., and Xiao, G.-Q.
\newblock Dual transformer for point cloud analysis.
\newblock \emph{IEEE Transactions on Multimedia}, 2022.

\bibitem[He et~al.(2022)He, Chen, Xie, Li, Doll{\'a}r, and
  Girshick]{he2022masked}
He, K., Chen, X., Xie, S., Li, Y., Doll{\'a}r, P., and Girshick, R.
\newblock Masked autoencoders are scalable vision learners.
\newblock In \emph{Proceedings of the IEEE/CVF Conference on Computer Vision
  and Pattern Recognition}, pp.\  16000--16009, 2022.

\bibitem[Huang et~al.(2021)Huang, Xie, Zhu, and Zhu]{huang2021spatio}
Huang, S., Xie, Y., Zhu, S.-C., and Zhu, Y.
\newblock Spatio-temporal self-supervised representation learning for 3d point
  clouds.
\newblock In \emph{Proceedings of the IEEE/CVF International Conference on
  Computer Vision}, pp.\  6535--6545, 2021.

\bibitem[Kratsios et~al.(2021)Kratsios, Zamanlooy, Liu, and
  Dokmani{\'c}]{kratsios2021universal}
Kratsios, A., Zamanlooy, B., Liu, T., and Dokmani{\'c}, I.
\newblock Universal approximation under constraints is possible with
  transformers.
\newblock \emph{arXiv preprint arXiv:2110.03303}, 2021.

\bibitem[Lai et~al.(2022)Lai, Liu, Jiang, Wang, Zhao, Liu, Qi, and
  Jia]{lai2022stratified}
Lai, X., Liu, J., Jiang, L., Wang, L., Zhao, H., Liu, S., Qi, X., and Jia, J.
\newblock Stratified transformer for 3d point cloud segmentation.
\newblock In \emph{Proceedings of the IEEE/CVF Conference on Computer Vision
  and Pattern Recognition}, pp.\  8500--8509, 2022.

\bibitem[Lee et~al.(2019)Lee, Lee, Kim, Kosiorek, Choi, and Teh]{lee2019set}
Lee, J., Lee, Y., Kim, J., Kosiorek, A., Choi, S., and Teh, Y.~W.
\newblock Set transformer: A framework for attention-based
  permutation-invariant neural networks.
\newblock In \emph{International conference on machine learning}, pp.\
  3744--3753. PMLR, 2019.

\bibitem[Li et~al.(2018)Li, Bu, Sun, Wu, Di, and Chen]{li2018pointcnn}
Li, Y., Bu, R., Sun, M., Wu, W., Di, X., and Chen, B.
\newblock Pointcnn: Convolution on x-transformed points.
\newblock \emph{Advances in neural information processing systems}, 31, 2018.

\bibitem[Liu et~al.(2019{\natexlab{a}})Liu, Han, Liu, and
  Zwicker]{liu2019point2sequence}
Liu, X., Han, Z., Liu, Y.-S., and Zwicker, M.
\newblock Point2sequence: Learning the shape representation of 3d point clouds
  with an attention-based sequence to sequence network.
\newblock In \emph{Proceedings of the AAAI Conference on Artificial
  Intelligence}, volume~33, pp.\  8778--8785, 2019{\natexlab{a}}.

\bibitem[Liu et~al.(2019{\natexlab{b}})Liu, Fan, Xiang, and
  Pan]{liu2019relation}
Liu, Y., Fan, B., Xiang, S., and Pan, C.
\newblock Relation-shape convolutional neural network for point cloud analysis.
\newblock In \emph{Proceedings of the IEEE/CVF Conference on Computer Vision
  and Pattern Recognition}, pp.\  8895--8904, 2019{\natexlab{b}}.

\bibitem[Liu et~al.(2021)Liu, Lin, Cao, Hu, Wei, Zhang, Lin, and
  Guo]{liu2021swin}
Liu, Z., Lin, Y., Cao, Y., Hu, H., Wei, Y., Zhang, Z., Lin, S., and Guo, B.
\newblock Swin transformer: Hierarchical vision transformer using shifted
  windows.
\newblock In \emph{Proceedings of the IEEE/CVF International Conference on
  Computer Vision}, pp.\  10012--10022, 2021.

\bibitem[Loshchilov \& Hutter(2016)Loshchilov and Hutter]{loshchilov2016sgdr}
Loshchilov, I. and Hutter, F.
\newblock Sgdr: Stochastic gradient descent with warm restarts.
\newblock \emph{arXiv preprint arXiv:1608.03983}, 2016.

\bibitem[Loshchilov \& Hutter(2017)Loshchilov and
  Hutter]{loshchilov2017decoupled}
Loshchilov, I. and Hutter, F.
\newblock Decoupled weight decay regularization.
\newblock \emph{arXiv preprint arXiv:1711.05101}, 2017.

\bibitem[Mao et~al.(2021)Mao, Xue, Niu, Bai, Feng, Liang, Xu, and
  Xu]{mao2021voxel}
Mao, J., Xue, Y., Niu, M., Bai, H., Feng, J., Liang, X., Xu, H., and Xu, C.
\newblock Voxel transformer for 3d object detection.
\newblock In \emph{Proceedings of the IEEE/CVF International Conference on
  Computer Vision}, pp.\  3164--3173, 2021.

\bibitem[Mazur \& Lempitsky(2021)Mazur and Lempitsky]{mazur2021cloud}
Mazur, K. and Lempitsky, V.
\newblock Cloud transformers: A universal approach to point cloud processing
  tasks.
\newblock In \emph{Proceedings of the IEEE/CVF International Conference on
  Computer Vision}, pp.\  10715--10724, 2021.

\bibitem[Metropolis \& Ulam(1949)Metropolis and Ulam]{metropolis1949monte}
Metropolis, N. and Ulam, S.
\newblock The monte carlo method.
\newblock \emph{Journal of the American statistical association}, 44\penalty0
  (247):\penalty0 335--341, 1949.

\bibitem[Misra et~al.(2021)Misra, Girdhar, and Joulin]{misra2021end}
Misra, I., Girdhar, R., and Joulin, A.
\newblock An end-to-end transformer model for 3d object detection.
\newblock In \emph{Proceedings of the IEEE/CVF International Conference on
  Computer Vision}, pp.\  2906--2917, 2021.

\bibitem[Pang et~al.(2022)Pang, Wang, Tay, Liu, Tian, and Yuan]{pang2022masked}
Pang, Y., Wang, W., Tay, F.~E., Liu, W., Tian, Y., and Yuan, L.
\newblock Masked autoencoders for point cloud self-supervised learning.
\newblock \emph{arXiv preprint arXiv:2203.06604}, 2022.

\bibitem[Qi et~al.(2017{\natexlab{a}})Qi, Su, Mo, and Guibas]{qi2017pointnet}
Qi, C.~R., Su, H., Mo, K., and Guibas, L.~J.
\newblock Pointnet: Deep learning on point sets for 3d classification and
  segmentation.
\newblock In \emph{Proceedings of the IEEE conference on computer vision and
  pattern recognition}, pp.\  652--660, 2017{\natexlab{a}}.

\bibitem[Qi et~al.(2017{\natexlab{b}})Qi, Yi, Su, and Guibas]{qi2017pointnet++}
Qi, C.~R., Yi, L., Su, H., and Guibas, L.~J.
\newblock Pointnet++: Deep hierarchical feature learning on point sets in a
  metric space.
\newblock \emph{Advances in neural information processing systems}, 30,
  2017{\natexlab{b}}.

\bibitem[Raffel et~al.(2020)Raffel, Shazeer, Roberts, Lee, Narang, Matena,
  Zhou, Li, Liu, et~al.]{raffel2020exploring}
Raffel, C., Shazeer, N., Roberts, A., Lee, K., Narang, S., Matena, M., Zhou,
  Y., Li, W., Liu, P.~J., et~al.
\newblock Exploring the limits of transfer learning with a unified text-to-text
  transformer.
\newblock \emph{J. Mach. Learn. Res.}, 21\penalty0 (140):\penalty0 1--67, 2020.

\bibitem[Ramachandran et~al.(2019)Ramachandran, Parmar, Vaswani, Bello,
  Levskaya, and Shlens]{ramachandran2019stand}
Ramachandran, P., Parmar, N., Vaswani, A., Bello, I., Levskaya, A., and Shlens,
  J.
\newblock Stand-alone self-attention in vision models.
\newblock \emph{Advances in Neural Information Processing Systems}, 32, 2019.

\bibitem[Sander et~al.(2022)Sander, Ablin, Blondel, and
  Peyr{\'e}]{sander2022sinkformers}
Sander, M.~E., Ablin, P., Blondel, M., and Peyr{\'e}, G.
\newblock Sinkformers: Transformers with doubly stochastic attention.
\newblock In \emph{International Conference on Artificial Intelligence and
  Statistics}, pp.\  3515--3530. PMLR, 2022.

\bibitem[Sharma \& Kaul(2020)Sharma and Kaul]{sharma2020self}
Sharma, C. and Kaul, M.
\newblock Self-supervised few-shot learning on point clouds.
\newblock \emph{Advances in Neural Information Processing Systems},
  33:\penalty0 7212--7221, 2020.

\bibitem[Shi et~al.(2021)Shi, Gao, Ren, Xu, Liang, Li, and
  Kwok]{shi2021sparsebert}
Shi, H., Gao, J., Ren, X., Xu, H., Liang, X., Li, Z., and Kwok, J. T.-Y.
\newblock Sparsebert: Rethinking the importance analysis in self-attention.
\newblock In \emph{International Conference on Machine Learning}, pp.\
  9547--9557. PMLR, 2021.

\bibitem[Srivastava et~al.(2014)Srivastava, Hinton, Krizhevsky, Sutskever, and
  Salakhutdinov]{srivastava2014dropout}
Srivastava, N., Hinton, G., Krizhevsky, A., Sutskever, I., and Salakhutdinov,
  R.
\newblock Dropout: a simple way to prevent neural networks from overfitting.
\newblock \emph{The journal of machine learning research}, 15\penalty0
  (1):\penalty0 1929--1958, 2014.

\bibitem[Thomas et~al.(2019)Thomas, Qi, Deschaud, Marcotegui, Goulette, and
  Guibas]{thomas2019kpconv}
Thomas, H., Qi, C.~R., Deschaud, J.-E., Marcotegui, B., Goulette, F., and
  Guibas, L.~J.
\newblock Kpconv: Flexible and deformable convolution for point clouds.
\newblock In \emph{Proceedings of the IEEE/CVF international conference on
  computer vision}, pp.\  6411--6420, 2019.

\bibitem[Touvron et~al.(2021)Touvron, Cord, Douze, Massa, Sablayrolles, and
  J{\'e}gou]{touvron2021training}
Touvron, H., Cord, M., Douze, M., Massa, F., Sablayrolles, A., and J{\'e}gou,
  H.
\newblock Training data-efficient image transformers \& distillation through
  attention.
\newblock In \emph{International Conference on Machine Learning}, pp.\
  10347--10357. PMLR, 2021.

\bibitem[Uy et~al.(2019)Uy, Pham, Hua, Nguyen, and Yeung]{uy2019revisiting}
Uy, M.~A., Pham, Q.-H., Hua, B.-S., Nguyen, T., and Yeung, S.-K.
\newblock Revisiting point cloud classification: A new benchmark dataset and
  classification model on real-world data.
\newblock In \emph{Proceedings of the IEEE/CVF international conference on
  computer vision}, pp.\  1588--1597, 2019.

\bibitem[Vaswani et~al.(2017)Vaswani, Shazeer, Parmar, Uszkoreit, Jones, Gomez,
  Kaiser, and Polosukhin]{vaswani2017attention}
Vaswani, A., Shazeer, N., Parmar, N., Uszkoreit, J., Jones, L., Gomez, A.~N.,
  Kaiser, {\L}., and Polosukhin, I.
\newblock Attention is all you need.
\newblock \emph{Advances in neural information processing systems}, 30, 2017.

\bibitem[Wang et~al.(2021)Wang, Liu, Yue, Lasenby, and
  Kusner]{wang2021unsupervised}
Wang, H., Liu, Q., Yue, X., Lasenby, J., and Kusner, M.~J.
\newblock Unsupervised point cloud pre-training via occlusion completion.
\newblock In \emph{Proceedings of the IEEE/CVF international conference on
  computer vision}, pp.\  9782--9792, 2021.

\bibitem[Wu et~al.(2015)Wu, Song, Khosla, Yu, Zhang, Tang, and Xiao]{wu20153d}
Wu, Z., Song, S., Khosla, A., Yu, F., Zhang, L., Tang, X., and Xiao, J.
\newblock 3d shapenets: A deep representation for volumetric shapes.
\newblock In \emph{Proceedings of the IEEE conference on computer vision and
  pattern recognition}, pp.\  1912--1920, 2015.

\bibitem[Xie et~al.(2018)Xie, Liu, Chen, and Tu]{xie2018attentional}
Xie, S., Liu, S., Chen, Z., and Tu, Z.
\newblock Attentional shapecontextnet for point cloud recognition.
\newblock In \emph{Proceedings of the IEEE conference on computer vision and
  pattern recognition}, pp.\  4606--4615, 2018.

\bibitem[Xu et~al.(2018)Xu, Fan, Xu, Zeng, and Qiao]{xu2018spidercnn}
Xu, Y., Fan, T., Xu, M., Zeng, L., and Qiao, Y.
\newblock Spidercnn: Deep learning on point sets with parameterized
  convolutional filters.
\newblock In \emph{Proceedings of the European Conference on Computer Vision
  (ECCV)}, pp.\  87--102, 2018.

\bibitem[Yan et~al.(2022)Yan, Yang, Li, Guan, Kang, Hua, and
  Huang]{yan2022implicit}
Yan, S., Yang, Z., Li, H., Guan, L., Kang, H., Hua, G., and Huang, Q.
\newblock Implicit autoencoder for point cloud self-supervised representation
  learning.
\newblock \emph{arXiv preprint arXiv:2201.00785}, 2022.

\bibitem[Yang et~al.(2019{\natexlab{a}})Yang, Zhang, Ni, Li, Liu, Zhou, and
  Tian]{yang2019modeling}
Yang, J., Zhang, Q., Ni, B., Li, L., Liu, J., Zhou, M., and Tian, Q.
\newblock Modeling point clouds with self-attention and gumbel subset sampling.
\newblock In \emph{Proceedings of the IEEE/CVF conference on computer vision
  and pattern recognition}, pp.\  3323--3332, 2019{\natexlab{a}}.

\bibitem[Yang et~al.(2019{\natexlab{b}})Yang, Dai, Yang, Carbonell,
  Salakhutdinov, and Le]{yang2019xlnet}
Yang, Z., Dai, Z., Yang, Y., Carbonell, J., Salakhutdinov, R.~R., and Le, Q.~V.
\newblock Xlnet: Generalized autoregressive pretraining for language
  understanding.
\newblock \emph{Advances in neural information processing systems}, 32,
  2019{\natexlab{b}}.

\bibitem[Yu et~al.(2022)Yu, Tang, Rao, Huang, Zhou, and Lu]{yu2022point}
Yu, X., Tang, L., Rao, Y., Huang, T., Zhou, J., and Lu, J.
\newblock Point-bert: Pre-training 3d point cloud transformers with masked
  point modeling.
\newblock In \emph{Proceedings of the IEEE/CVF Conference on Computer Vision
  and Pattern Recognition}, pp.\  19313--19322, 2022.

\bibitem[Yun et~al.(2019)Yun, Bhojanapalli, Rawat, Reddi, and
  Kumar]{yun2019transformers}
Yun, C., Bhojanapalli, S., Rawat, A.~S., Reddi, S.~J., and Kumar, S.
\newblock Are transformers universal approximators of sequence-to-sequence
  functions?
\newblock \emph{arXiv preprint arXiv:1912.10077}, 2019.

\bibitem[Yun et~al.(2020)Yun, Chang, Bhojanapalli, Rawat, Reddi, and
  Kumar]{yun2020n}
Yun, C., Chang, Y.-W., Bhojanapalli, S., Rawat, A.~S., Reddi, S., and Kumar, S.
\newblock O (n) connections are expressive enough: Universal approximability of
  sparse transformers.
\newblock \emph{Advances in Neural Information Processing Systems},
  33:\penalty0 13783--13794, 2020.

\bibitem[Zaheer et~al.(2017)Zaheer, Kottur, Ravanbakhsh, Poczos, Salakhutdinov,
  and Smola]{zaheer2017deep}
Zaheer, M., Kottur, S., Ravanbakhsh, S., Poczos, B., Salakhutdinov, R.~R., and
  Smola, A.~J.
\newblock Deep sets.
\newblock \emph{Advances in neural information processing systems}, 30, 2017.

\bibitem[Zaheer et~al.(2020)Zaheer, Guruganesh, Dubey, Ainslie, Alberti,
  Ontanon, Pham, Ravula, Wang, Yang, et~al.]{zaheer2020big}
Zaheer, M., Guruganesh, G., Dubey, K.~A., Ainslie, J., Alberti, C., Ontanon,
  S., Pham, P., Ravula, A., Wang, Q., Yang, L., et~al.
\newblock Big bird: Transformers for longer sequences.
\newblock \emph{Advances in Neural Information Processing Systems},
  33:\penalty0 17283--17297, 2020.

\bibitem[Zhang et~al.(2021)Zhang, Wan, Shen, and Wu]{zhang2021pvt}
Zhang, C., Wan, H., Shen, X., and Wu, Z.
\newblock Pvt: Point-voxel transformer for point cloud learning.
\newblock \emph{arXiv preprint arXiv:2108.06076}, 2021.

\bibitem[Zhao et~al.(2021)Zhao, Jiang, Jia, Torr, and Koltun]{zhao2021point}
Zhao, H., Jiang, L., Jia, J., Torr, P.~H., and Koltun, V.
\newblock Point transformer.
\newblock In \emph{Proceedings of the IEEE/CVF International Conference on
  Computer Vision}, pp.\  16259--16268, 2021.

\end{thebibliography}
\bibliographystyle{icml2023}

\newpage
\appendix
\onecolumn

\section{Notations} \label{app:notation}
\bgroup
\def\arraystretch{1.4}
\begin{tabular}{p{1in}p{4in}}
$\displaystyle f$ & a continuous function\\
$\displaystyle g$ & transformer\\
$\displaystyle \overline{g}$ & modified transformer\\
$\displaystyle \mathcal{F}$ & the class of continuous sequence-to-sequence function\\
$\displaystyle \mathcal{F}_S$ & the class of continuous set-to-set function\\
$\displaystyle \overline{\mathcal{F}}$ & the class of piece-wise constant sequence-to-sequence function\\
$\displaystyle \overline{\mathcal{F}}_S$ & the class of piece-wise constant set-to-set function\\
$\displaystyle \mathcal{T}^{h,m,r}$ & the class of (sparse) transformers with $h$ attention heads, $m$ head size, and hidden layer width $r$\\
$\displaystyle \overline{\mathcal{T}}^{h,m,r}$ & the class of the modified transformers with $h$ attention heads, $m$ head size, and hidden layer width $r$\\
$\displaystyle \sigma_S$ & softmax activation\\
$\displaystyle \sigma_H$ & hardmax activation\\
$\displaystyle \ell_p$ & p norm\\
\\
$\displaystyle \mathbb{G}_\delta$ & grid $\{0, \delta, \dotso, 1-\delta\}^{d\times n}$\\
$\displaystyle \mathbb{G}^{+}_\delta$ & extend grid $\{-\delta^{-nd}, 0, \delta, \dotso, 1-\delta\}^{d\times n}$\\
\\
$\displaystyle n$ & number of points/elements/tokens\\
$\displaystyle d$ & point/element/token feature size\\
$\displaystyle m$ & head size\\
$\displaystyle h$ & heads number\\
$\displaystyle r$ & hidden layer width\\
$\displaystyle \delta$ & step size\\
\\
$\displaystyle \bm X$ & transformer input\\
$\displaystyle \subX^i$ & $i$-th subset of transformer input\\
$\displaystyle \bm P$ & $xyz$ coordinates for point cloud (set)\\
$\displaystyle \bm E$ & positional embedding\\
$\displaystyle \bm L$ & quantized transformer input\\
$\displaystyle \bm A_{\bm L}$ & desired output for the input $\bm L$\\
$\displaystyle \bm W_V^i$ & value parameter in $i$-th single-head attention layer\\
$\displaystyle \bm W_K^i$ & key parameter in $i$-th single-head attention layer\\
$\displaystyle \bm W_Q^i$ & query parameter in $i$-th single-head attention layer\\
$\displaystyle \bm W_O$ & multi-head attention parameter\\
$\displaystyle \bm W_1$ & feed-forward layer parameter\\
$\displaystyle \bm W_2$ & feed-forward layer parameter\\
$\displaystyle \bm W_p$ & parameter for position embedding\\
\end{tabular}
\egroup

\bgroup
\def\arraystretch{1.4}
\begin{tabular}{p{1in}p{4in}}
\\
$\displaystyle \bm u$ & query, key, and value parameter used in universal approximation proof\\
$\displaystyle \bm e^{(1)}$ & indicator vector $(1,0,0,\dotso,0)\in \mathbb{R}^d$\\
$\displaystyle \bm 1_n$ & vector with all ones $(1,\dotso,1)\in \mathbb{R}^n$\\
$\displaystyle \bm 0_n$ & vector with all zeros $(0,\dotso,0)\in \mathbb{R}^n$\\
\\
$\displaystyle \text{Head}^i(\cdot)$ & $i$-th single-head attention layer\\
$\displaystyle \text{SHead}^i(\cdot)$ & $i$-th sparse/sampled single-head attention layer\\
$\displaystyle \text{Attn}(\cdot)$ & multi-head attention layer\\
$\displaystyle \text{SAttn}(\cdot)$ & multi-head attention layer with sampled sparse attention\\
$\displaystyle \text{TB}(\cdot)$ & transformer block\\
$\displaystyle \text{STB}(\cdot)$ & sampled transformer block\\
$\displaystyle t(\cdot)$ & a series of any number of transformer blocks\\
$\displaystyle q_c(\cdot)$ & contextual mapping\\
$\displaystyle \bm \Psi(\cdot;b_Q, b'_Q)$ & selective shift operation\\
$\displaystyle \psi(\cdot;b_Q)$ & a single-head attention in selective shift operation\\
$\displaystyle d_c(\cdot, \cdot)$ & distance between two functions\\
\end{tabular}
\egroup

\section{Additional Information on the Basic Classification Setting}
\subsection{$k$NN Transformer}\label{app:ua_spth}
\begin{definition}[$k$NN Attention]\label{def:knn_atten}
    For $k \in [n]$, $k$NN attention has the attention pattern $\mathcal{A}_k = \text{kNN}(k)$ for all points, where $\text{kNN}(\cdot)$ represents the Euclidean $k$-nearest neighbourhood of the input.
\end{definition} 

\begin{definition}[$k$NN Transformer]\label{def:knn_tf}
    The $k$NN transformer is the transformer defined as in Eq.~\ref{eq:dense_tf}, but with the $k$NN attention of definition~\ref{def:knn_atten}.
\end{definition} 



In addition, in the case of vector attention (Eq. 3 in~\citep{zhao2021point}), universal approximation holds as the learnable mapping $\gamma(\cdot)$ (an MLP) is a universal approximator.
This may helps to explain why vector attention could outperform scalar attention in Tab.~7 of \citep{zhao2021point}.

Finaly, in Tab.~\ref{tab:single_layer}, the performance of the $k$NN transformer drops with the increasing number of points.
This is because as the point number increase, the fix $k$ nearest neighbor number is relatively reduced.
As a result, the receptive field shrink.
So the performance drops.

\subsection{Memory Usage}\label{app:memory_usage}
\begin{table*}[t]
    \caption{%
        Memory usage (Gb) for different attention mechanisms in the basic classification setting.
        All are trained on a single RTX 3090 with 24 Gb on board RAM.
        OM denotes \textit{out of memory}.%
    }%
    \label{tab:memory}%
    \centering
        \resizebox{0.9\textwidth}{!}{%
        \begin{tabular}{l|cccccccc}
            \toprule
            \#Points & 256 & 512 & 768 & 1024 & 2048 & 3072 & 4096 & 8192 \\
            \midrule
            MLP + FC (no transformer) & 0.9 & 0.9 & 0.9 & 1.0 & 1.1 & 1.1 & 1.2  & 1.5 \\
            Dense Attention & 1.2 & 1.8 & 2.7 & 3.9 & 11.9 & OM & OM  & OM \\
            Sparse Attention & 1.0 & 1.1 & 1.2 & 1.3 & 1.7 & 2.1 & 2.5  & 4.3 \\
            Sampled Attention & 1.0 & 1.1 & 1.2 & 1.3 & 1.7 & 2.1 & 2.5  & 4.2 \\
            $k$NN Attention & 1.9 & 2.8 & 3.7 & 4.4 & 8.5 & 11.4 & 16.5 & OM \\
            \bottomrule
        \end{tabular}
        }
\end{table*}
The memory usage of some sparse attentions in basic setting is in Tab~\ref{tab:memory}, which shows that the dense transformer has the largest memory usage due to its $O(n^2)$ complexity.
The sparse transformer and sampled transformer have comparable memory usage due to the same $O(n)$ complexity.

\subsection{In comparison with Inducting Points (Set Transformer)}\label{app:inducting}
We additionally compared the proposed sampled attention with learnable inducting points strategy~\citep{lee2019set}.
The inducting points here are implemented by simply replacing the multi-heads self-attention transformer block in Eq. \ref{seq:stacked_sampled_attn} with the Induced Set Attention Block (ISAB) in Eq. (9) of \citet{lee2019set}.
And the positional embedding is added in the key and value input as per our sampled attention.
Our implementation of the basic classification in Sec.~\ref{subsec:basic_setting} is different from the one in \citet{lee2019set} with respect to the data pre-processing: our data pre-processing is in line with \citet{zhao2021point,yu2022point}, while \citet{lee2019set} follow \citet{zaheer2017deep} without positional embedding.
As we can see in Tab.~\ref{tab:single_layer}, our proposed sampled attention outperformance the inducting point strategy~\citep{lee2019set} with linear complexity in the attention matrix.

As the performance of \citet{lee2019set} on the two implements is quite different, we further compared the sampled attention and inducting points strategy in the implementation provided by 
\hyperlink{https://github.com/juho-lee/set_transformer}{the official implementation of \citet{lee2019set}}.
To begin with, our proposed sampled attention could be applied to the inducting points strategy directly to reduce its complexity from $O(mn)$ to $O(n)$, where $n$ is the input points number and $m$ is the learnable inducting points number.
Specifically, we use the sampled attention to replace the dense attention in the Induced Set Attention Block(ISAB) from the Eq. 9 of \citet{lee2019set}.
However, as the inducting points and points have different physical meanings, also as the inducting points number $m$ (query in the self-attention) is not equal to the input points number $n$ (key and value), our Hamiltonian cycle attention could not be applied directly. 
We instead applied a different version of sampled attention by randomly sampling two elements per row in the dense attention matrix. 
This is a loose version of sampled attention as no Hamiltonian cycle is constructed.
The results could be found in Tab.~\ref{tab:class_inducting}.
As we can see, our proposed sampled attention is still comparable with the set transformer but with less computational complexity.

\begin{table}[t]
    \caption{Object classification in the setting of \cite{lee2019set} measured accuracy (\%).}%
    \label{tab:class_inducting}%
    \centering
        \resizebox{0.8\textwidth}{!}{%
        \begin{tabular}{l|cccccccc}
            \toprule
            \#Points & 100 & 200 & 1000 & 2000 & 3000 & 5000 \\
            \midrule
            ISAB(16) + PMA & 80.52 & 85.38 & 84.43 & 85.99 & 85.49 & 86.99 \\
            ISAB(16) + PMA + sampled attention (ours) & 81.25 & 82.65 & 84.15 & 85.04 & 84.48 & 86.49 \\
            \bottomrule
        \end{tabular}
        }
\end{table}
\subsection{In comparison with Stratified Strategy}\label{app:stratified}
The window-based transformer is another important branch of exploring the representation power of the transformer. 
Combined with the hierarchical backbone, it has been widely used in processing 2D images, languages, and 3D point clouds, such as \citet{liu2021swin,lai2022stratified}.
The window-based transformer is proposed to learn the cross-window relationships as well as the non-overlapping local relationship.

Here we compared our proposed sampled attention with the Stratified strategy from Figure 3 of \citet{lai2022stratified} in Tab.~\ref{tab:single_layer}.
The Stratified strategy could be viewed as a combination of dense and sparse keys obtained by the window partition of different sizes.
It is an efficient design for learning token relationships in the hierarchical backbone. However, in the single-layer setting, directly learning $O(n^2)$ connections in the attention matrix may be a better solution as it could reach the full receptive field.
As our proposed sampled attention mechanism could estimate $O(n^2)$ connections by implementing the Monto Carlo simulation,
we outperformed the Stratified strategy in the basic classification setting as per Tab.~\ref{tab:single_layer}.

\section{Amortized Clustering with Mixture of Gaussians}\label{app:clustering}
We additionally tested the proposed sampled attention in 2D set datasets in the encoding-decoding framework introduced by \citet{lee2019set}. And the task is about 
using a neural network to learn the parameters of the mixture Gaussian distribution from the input set data.

To begin with, the mixture Gaussian distribution is defined by a weighted sum of $k$ number of Gaussian distribution.
Given a dataset $\bm X=\{ \bm x_1, \dotso, \bm x_n \}$, the log-likelihood of the mixture Gaussian distribution is defined as follows:
\begin{align}\label{eq:log_likelihood}
    \log p(\bm X; \bm \theta) = \sum_{i=1}^n \log \sum_{j=1}^k \pi_j \mathcal{N}(\bm x_i; \bm \mu_i; \text{diag}(\bm \sigma_j^2)).
\end{align}

Generally, the parameters of the mixture Gaussian distribution are inferred by maximizing the log-likelihood $\theta^*(\bm X) = \argmax_\theta \log p(\bm X; \theta)$ using Expectation-Maximisation (EM) algorithm as the closed-form solution could not be inferred directly by setting the gradient equals to zero.
Here we instead use the transformer to infer $\theta^*(\bm X)$.
Specifically, given the input, the neural network $f$ outputs mixture Gaussian parameters $f(\bm X)=\{ \pi(\bm X), \{ \mu_j(\bm x), \sigma_j(\bm X) \}_{j=1}^k \}$ by maximing the log likelihood in Eq.~\ref{eq:log_likelihood} (and replacing all parameters as functions of $\bm X$).

The 2D set data $\bm X$ is randomly sampled from a given mixture Gaussian distribution with $k=4$. And the number of elements $n$ is randomly sampled from $[100, 500]$. Namely, when setting the dimension of Gaussian distribution as 2, each sampled point could be viewed as a 2D data point, so the sampled collection is a 2D set dataset.

The baseline we compared with is the Set transformer~\citep{lee2019set} with two Induced Set Attention Block(ISAB) in the encoder, one Multi-head Attention (PMA) and two Set Attention Block (SAB) in the decoder, as per \href{https://github.com/juho-lee/set_transformer}{the official implementation}.
The inducting points refer to the additional learnable points $\bm I \in \mathbb{R}^{m\times d}$ proposed in Eq. 9 of \citep{lee2019set}, with $d$ dimension and $m$ number of inducting points.
Here we have a mixture usage of points, tokens, and elements to represent a single sampled data point $x_i$.

As the computation complexity of the inducting points block (ISAB) is $O(nm)$, our sampled attention may be adopted in the ISAB to reduce the computation complexity to $O(n)$.
However, as the number of inducting points $m$ (regarded as the query in \citet{lee2019set}) is not equal to the number of input points $n$ (regarded as key and value) (in fact inducting points and points have different physical meanings), our Hamiltonian cycle attention could not be applied directly.
In fact, the dense attention matrix in the inducting points layer is $m\times n$ rather than $n\times n$.
We instead applied a different version of sampled attention by randomly sampling two elements per row in the attention matrix.
This is a loose version of sampled attention as no Hamiltonian cycle is constructed.
As we can see in Tab.~\ref{tab:clustering}, the sampled attention could be a plug-in module to replace the dense attention in the inducting points structure with competitive performance but theoretically less computational complexity.

\begin{table}[t]
    \caption{%
        Amortized clustering results. The number in ISAB($\cdot$) indicates the number of learnable inducting points used in ISAB as per \citet{lee2019set}.
        The evaluation metric LLO/data is the average log-likelihood value, and LL1/data is the average log-likelihood value after a single EM update (implemented by scikit-learn package \citep{sklearn_api}).
    }%
    \label{tab:clustering}%
    \centering
        \resizebox{0.7\textwidth}{!}{%
        \begin{tabular}{l|ccc}
            \toprule
            Architecture & LL0/data & LL1/data \\
            \midrule
            rFF + Pooling & -2.0006 $\pm$ 0.0123 & -1.6186 $\pm$ 0.0042  \\
            ISAB(16) + PMA & -1.5034 $\pm$ 0.0072 & -1.4908 $\pm$ 0.0044 \\
            \midrule
            ISAB(16) + PMA + sampled attention (ours) & -1.5663 $\pm$ 0.0074 & -1.5272 $\pm$ 0.0052 \\
            \bottomrule
        \end{tabular}
        }
\end{table}

\section{Inductive Bias}\label{app:inductive_bias}
We use inductive bias to refer to the prior knowledge and design built into a machine learning model.
Loosely, more inductive bias may have better performance in specific tasks, while less inductive bias may have better generalisation ability (meaning, for example, wider applicability to different tasks and frameworks), and fewer hyperparameters that need to be tuned.

In this paper, our initial research goal is to have an efficient and permutation invariant transformer for point sets / clouds.
Both nearest neighbour search and inducing points are good designs as both models are efficient and permutation invariant.
However, to implement the nearest neighbour search, one should introduce the hyperparameter of the number of neighbours, and the choice of definition of token-to-token distance.
Further, the inducing points introduced additional parameters (inducing points themselves), which means additional backpropagation calculations.
Close inspection by \citet{lee2019set} reveals a number of other non-trivial design choices.
In contrast to e.g. the nearest neighbour based approaches, our random permutation-based attention involves less intuition-guided assumptions and fewer additional hyper parameter choices.

\section{Universal Approximator Proof}\label{app:ua}
A proof of Corollary~\ref{thm:our_ua} follows the steps described in \S~\ref{subsec:ua}. 
As we only changed the dense/sparse attention to the sampled attention, 
the steps \ding{172} and \ding{174} in \S~\ref{subsec:ua} remain the same as \cite{yun2019transformers,yun2020n} and found in the \S C and F in \cite{yun2020n}.
Here we need only cover the proof of step \ding{173}.

First, we have $\mathcal{F}_S(\cdot)$ is the class of continuous set-to-set function, and $\overline{\mathcal{F}}_S(\cdot)$ is the class of piece-wise constant set-to-set function.

\begin{lemma}[Modified Universal Approximation.]\label{prop:modified_ua}
  For each $\overline{f} \in \overline{\mathcal{F}}_S(\delta)$ and $1 \leq q < \infty$, $\exists \overline{g} \in \overline{\mathcal{T}}^{2,1,1}$ such that $\overline{f}(\bm X)=\overline{g}(\bm X)$ for all $\bm X \in \mathbb{D}$.
\end{lemma}
Without loss of generality, here $\mathbb{D}\in [0, 1)^{d\times n}$.
As in \citep{yun2019transformers, yun2020n} The proof of Lemma~\ref{prop:modified_ua} could then be separated into four steps:
\begin{enumerate}
   \item Use the positional embedding $\bm E$ in \S~\ref{subsec:tf} such that each column of the input $\bm X_k + \bm E_k$ are in disjoint intervals.
   \item The input $\bm X + \bm E$ is quantized into $\bm L$ with values in $\{0, \delta, \dotso, n-\delta\}$ by a series of modified feed-forward layers.
   \item The \textit{contextual mapping q} defined in Definition~\ref{def:contextual_mapping} is implemented by a series of modified sampled multi-head self-attention layers (modified version of Eq.~\ref{eq:sampled_attn}) with the input of $\bm L$ .
   \item Another series of modified feed-forward layers implements the \textit{value mapping} such that each element in the unique id $q(\bm L)$ is mapped to the desired output $\bm A_{\bm X}$.
\end{enumerate}
As modified feed-forward layers are all the same as in \cite{yun2020n}, the definition and proof of step 2 is available in \S D.2 and E.1 in \citep{yun2020n}, while the definition and proof of step 4 could be found in the \S D.4 and E.3 in \citep{yun2020n}.
Here we mainly explain steps 1 and 3.

\subsection{Positional Embedding}
The positional input for point sets in its $xyz$ coordinate $\bm P \in \R^{3\times n}$.
We adopted a matrix $\bm W_p\in \R^{d\times 3}$ (a permutation invariant operation) such that the input of the sampled transformer will be $\bm X + \bm E = \bm X + \bm W_p {\bm P}$. And there exists a case such that:
\begin{align}
   \bm E_1 = (n-1)\bm 1_n, \text{ and } \bm E = (i-2)\bm 1_n, \text{ for } i \in [2:n].
\end{align}
In this case, the first column will be $(\bm X + \bm E)_1 \in [n-1, n)^d$, and $(\bm X + \bm E)_i \in [i-2, i-1)^d$ for $i\in [2:n]$.
So the requirement of step 1 is satisfied, that each column lies in disjoint intervals.

\subsection{Contextual Mapping for Stacked Multi-Heads self-Attention Layers}
After the step 2, the quantized input $\bm L$ will be in the set $\mathbb{H}_\delta \subset \R^{d\times n}$, such that:
\begin{align}
   \mathbb{H}_\delta := \{ \bm G + \bm E \in \R^{d\times n} \vert \bm G \in \mathbb{G}_\delta\},
\end{align}
with $\mathbb{G}_\delta := \{0, \delta, \dotso, 1-\delta\}$.
Then the adaptive selective shift operation $\Psi$ is defined so that the learnable parameter $\bm u^T \in \R^{d}$ could map $\bm u^T\Psi(\bm L)$ into unique scalars (ids).
Finally, with the help of the all-max-shift operation $\Omega$, the output of a series of those two operations will be a scalar in disjoint intervals w.r.t each column of $\bm L$, as well as different inputs $\bm L$ and $\bm L'$, thereby implementing the contextual mapping in Definition.~\ref{def:contextual_mapping}.

\paragraph{Adaptive Selective Shift Operation.} With a 2 heads and 1 hidden layer width modified multi-heads attention layer, the adaptive selective shift operation $\Psi(\cdot)$ may be defined as:
\begin{subequations}
    \begin{align}
       \Psi^l(\bm L^l;c, b_Q, b'_Q) :&= \bm L^l + c[\bm 1^1_{n_s} -\bm 1^1_{n_s}] 
       \begin{bmatrix}
          \psi^l(\bm L^l; b_Q) \\
          \psi^l(\bm L^l; b'_Q)
       \end{bmatrix}\\\label{eq:selective_op}
       \psi^l(\bm L^l; b_Q)_k &= \bm u^T \bm L_{\mathcal{A}_k^l} \sigma_H \left[(\bm u^T \bm L_{\mathcal{A}_k^l})^T(\bm u^T \bm L^l_k -b_Q)\right] \nonumber\\
       &= 
       \begin{cases}
          \max_{j\in \mathcal{A}^l_k} \bm u^T \bm L^l_j &\text{if } \bm u^T \bm L^l_k > b_Q\\
          \min_{j\in \mathcal{A}^l_k} \bm u^T \bm L^l_j &\text{if } \bm u^T \bm L^l_k < b_Q,
       \end{cases}
    \end{align}
 \end{subequations}
where we assign query, key, and value parameters as $\bm u^T$,
and we introduced the superscript $l$ to denote different attention layers of self-attention layer $l$.
With the help of hardmax, the $k$-th row of the attention matrix will be one-hot vectors to select the max or min vector in $\mathcal{A}^l_k$.
$\bm W_O=c[\bm 1^1_{n_s} -\bm 1^1_{n_s}] \in \mathbb{R}^{n_s\times 2}$ is used to make sure only the first element in feature dimension are changed in selective shift operation.
Specifically, the $1,k$-entity of the self-attention output reads:
\begin{align}
   \Psi^l(\bm L^l; c, b_Q, b'_Q)_{1,k} &= \overline{L}^l_{1,k} + c\left( \psi^l(\bm L^l; b_Q)_k - \psi^l(\bm L^l; b'_Q)_k \right)\\
   &=
   \begin{cases}
    \overline{L}^l_{1,k} + c\left( \max_{j\in \mathcal{A}^l_k} \bm u^T \bm L^l_j - \min_{j\in \mathcal{A}^l_k} \bm u^T \bm L^l_j  \right) & \text{if } b_Q < \bm u^T \bm L^l_k < b'_Q,\\
    \overline{L}^l_{1,k} & \text{if } \bm u^T \bm L^l_k \notin [b_Q, b'_Q].
   \end{cases}
\end{align}


Without loss of generality, the sampled transformer in \S.~\ref{sec:sampled_attention} may be viewed as a series of stacked masked attention $\mathcal{A}^i$ for $i\in [n]$, such that:
\begin{subequations}\label{eq:selective_attn}
\begin{align}
   &\mathcal{A}^i_{k =1+(i-2+n \mod n)} = \{ i, 1+ (i-2 + n \mod n) \} \\
   &\mathcal{A}^i_{k =i} = \{ i \} \\
   &\mathcal{A}^i_{k \not\subset\{ i,i-1 \mod n \}} = \{\},
\end{align}
\end{subequations}
for $k\in [n]$. This is in fact the $n$ point pairs in the Hamiltonian cycle. So the stack of all the masked attention is the cycle attention in Fig.~\ref{fig:cycle} reflected across the diagonal line. Then the Eq.~\ref{seq:stacked_sampled_attn} will be
\begin{align}
   \text{SAttn} (\bm L) = g^n (\bm L_{\mathcal{A}_n}) \circ g^{n-1} (\bm L_{\mathcal{A}_{n-1}}) \circ \cdots \circ g^1 (\bm L_{\mathcal{A}_1}),
\end{align}
noting that the updated column for previous $g^i$ will be applied to the next $g^{i+1}$.
In conclusion, the contextual mapping holds as the masked attention $\mathcal{A}^i$ is designed to aggregate information from all $n$ elements / tokens by applying the $g(\cdot)$ about $O(n)$ times, which matches the design of \citep{yun2020n}.

Now consider $\bm u^T = (1, \delta^{-1}, \delta^{-2}, \dotso, \delta^{-d+1})$, the mapping $l_i = p_s({ \bm L}_i) = \bm u^T \bm L_i$ is bijective  as all input point features $\bm L_i$ are different with at least one element having a gap of $\delta$.
In addition, without loss of generality, the order $l_2<l_3<\dotso<l_n<l_1$ holds as in \citep{yun2020n} because of the positional embedding $\bm E$.
Further, as each $l_i$ has $\delta^{-d}$ intervals, and as the $n$ tokens are disjoint with each other, we need $n\delta^-d$ adaptive selective operations to achieve the bijective mapping of unique ids.

\paragraph{First $\delta^{-d}$ selective shift operations.}
The first $\delta^{-d}$ layers are all applied to the second column (token) within $l_2 \in \left[ 0: \delta: \delta^{-d+1} - \delta \right]$,
and each selective shift operation will match one interval within $b_Q = b-\frac{\delta}{2}, b'_Q = b+\frac{\delta}{2}$ for $b \in \left[ 0: \delta: \delta^{-d+1} - \delta \right]$.
Also $\mathcal{A}^2$ is in fact $\mathcal{A}^2_1 = \{ 1 \}$, $\mathcal{A}^2_2 = \{ 1, 2 \}$, and is empty otherwise.
So all $\delta^{-d}$ layers are only applied on the first two token embeddings, then the maximum value is $l_1$ and the minimum value is $l_2$.
We have the output after those selective shift operations:
\begin{align}
   \tilde{l}_2 = l_2 + \delta^{-d} (\max_{j\in \mathcal{A}^1_2} l_j - \min_{j\in \mathcal{A}^1_2} l_j) = l_2 + \delta^{-d} (l_1 - l_2),
\end{align}
where with constant value $c=\delta^{-d}$ in Eq.~\ref{eq:selective_op}. Note that $\tilde{l}_2 > l_1$ because
\begin{align}
   l_2 + \delta^{-d} (l_1 - l_2) > l_1 \Leftrightarrow (\delta^{-d} - 1)(l_1 - l_2) > 0,
\end{align}
which is true.
So the current order becomes $l_3 < l_4 < \dotso < l_n < l_1 < \tilde{l}_2$.
So in the next $\delta^{-d}$ selective shift operations, the maximum value will be $\tilde{l}_2$ and the minimum will be $l_3$.

\paragraph{Second $\delta^{-d}$ selective shift operations.}
The next $\delta^{-d}$ layers will be applied on the third column (token embedding) within intervals $l_3 \in \left[ \sum_{i=0}^{d-1} \delta^{-i}: \delta: \sum_{i=0}^{d-1} \delta^{-i} + \delta^{-d+1} - \delta \right]$ which results in
\begin{align}
   \tilde{l}_3 = l_3 + \delta^{-d}(\tilde{l}_2-l_3) = l_3 + \delta^{-d}(l_2-l_3) + \delta^{-2d}(l_1-l_2),
\end{align}
which is again $\tilde{l}_3 > \tilde{l}_2$ because
\begin{align}
   l_3 + \delta^{-d} (\tilde{l}_2 - l_3) > \tilde{l}_2 \Leftrightarrow (\delta^{-d} - 1)(\tilde{l}_2 - l_3) > 0.
\end{align}
So we have a new maximum $\tilde{l}_3$ and new minimum $l_4$.

\paragraph{Repeat after $(n-1)\delta^{-d}$ operations.}
The next $\delta^{-d}$ will operate on the fourth column.
After all $(n-1)\delta^{-d}$ operations we have
\begin{align}
   (n-1)\sum_{i=0}^{d-1}\delta^{-i} \leq l_1 < \tilde{l}_2 < \dotso < \tilde{l}_n.
\end{align}
For $j$-th column, we will have the output
\begin{subequations}
   \begin{align}
      &\tilde{l}_1 = l_1,\\
      &\tilde{l}_2 = l_2 + \delta^{-d}(l_1 - l_2),\\
      &\tilde{l}_j = l_j + \sum^{j-2}_{k=1}\delta^{-kd} (l_{j-k}-l_{j-k+1}) + \delta^{-(j-1)d}(l_1 - l_2).
   \end{align}
\end{subequations}
And we also know the interval of each $l_i$
\begin{align}
   &l_1 \in [(n-1)\Delta:\delta:(n-1)\Delta+\delta^{-d+1}-\delta]\\
   &l_i \in [(i-2)\Delta:\delta:(i-2)\Delta+\delta^{-d+1}-\delta],
\end{align}
with $\delta^{-d+1}-\delta<\Delta:=\sum_{i=0}^{d-1}\delta^{-i}=\frac{\delta^{-d}-1}{\delta^{-1}-1} \leq \delta^{-d}-1 \Rightarrow 0 < \delta \leq \frac{1}{2}$.
So we have
\begin{align}
   &l_1 - l_2 \in [(n-1)\Delta - \delta^{-d+1} + \delta:\delta:(n-1)\Delta + \delta^{-d+1} - \delta]\\
   &l_i - l_{i+1} \in [-\Delta-\delta^{-d+1}+\delta:\delta:-\Delta+\delta^{-d+1}-\delta] \text{ for } i \in \{ 2, 3, \dotso, n-1 \}.
\end{align}
Then the interval of outputs are
\begin{align}
   &\tilde{l}_1 \in [(n-1)\Delta, (n-1)\Delta+\delta^{-d+1}-\delta]\\
   &\tilde{l}_2 \in [(n-1)\Delta \delta^{-d}-\delta^{-2d+1}+\delta^{-d+1}, (n-1)\Delta \delta^{-d}+\delta^{-2d+1}-\delta]\\
   &\tilde{l}_i \in [(i-2)\Delta - \sum_{k=1}^{i-2}\delta^{-kd} \Delta - \sum_{k=1}^{i-2}\delta^{-kd} (\delta^{-d+1}-\delta)+\delta^{-(i-1)d}(n-1)\Delta-\delta^{-(i-1)d}(\delta^{-d+1}-\delta),\nonumber \\
   & \qquad (i-2)\Delta +\delta^{-d+1}-\delta - \sum_{k=1}^{i-2}\delta^{-kd} \Delta \nonumber\\
   & \qquad \qquad+ \sum_{k=1}^{i-2}\delta^{-kd}(\delta^{-d+1}-\delta ) + \delta^{-(i-1)d}(n-1)\Delta+\delta^{-(i-1)d}(\delta^{-d+1}-\delta) ],
\end{align}
and to check whether intervals are disjoint or not, we take the difference between the lower bound of $\tilde{l}_{i+1}$ and the upper bound of $\tilde{l}_i$
\begin{align}
   \tilde{l}_{i+1}^{l} - \tilde{l}_i^{u} &= \Delta - \delta^{-(i-1)d} \Delta + (\delta^{-id} - \delta^{-(i-1)d})(n-1)\Delta - (\delta^{-d+1}-\delta) \\
   &\quad - \delta^{-(i-1)d}(\delta^{-d+1}-\delta) - 2\sum_{k=1}^{i-2}\delta^{-kd}(\delta^{-d+1}-\delta)\\
   &\quad - \delta^{-id}(\delta^{-d+1}-\delta) - \delta^{-(i-1)d}(\delta^{-d+1}-\delta) \\
   &= \left[ 1 - n \delta^{-(i-1)d} + (n-1)\delta^{-id} \right] \Delta \nonumber\\
   &\quad- \left( \frac{1+\delta^{-d}}{1-\delta^{-d}} - \frac{2\delta^{-d}}{1-\delta^{-d}} \delta^{-(i-2)d} + 2\delta^{-(i-1)d} + \delta^{-id} \right)(\delta^{-d+1}-\delta)\\
   &\geq \left[ \frac{2\delta^{-d}}{\delta^{-d}-1} - \frac{2\delta^{-d}}{\delta^{-d}-1}\delta^{-(i-2)d} - (n+2)\delta^{-(i-1)d} + (n-2)\delta^{-id}  \right] (\delta^{-d+1}-\delta)\\
   &\geq \delta^{-(i-2)d} \left[ -\frac{2\delta^{-d}}{\delta^{-d}-1} - (n+2)\delta^{-d} + (n-2)\delta^{-2d} \right] (\delta^{-d+1}-\delta)\\
   &\geq \delta^{-(i-2)d} \left[ -4- (n+2)\delta^{-d} + (n-2)\delta^{-2d} \right] (\delta^{-d+1}-\delta),
\end{align}
which is not guaranteed to be above 0, so the addition operations should be introduced.

Further, the adaptive shift operation is a one-to-one map as the map $\bm L_k \mapsto \bm u^T \bm L_k$ is one-to-one, and the permutation of columns is one-to-one, and so it sufficies to prove that the map $[l_1 \cdots l_n] \mapsto \tilde{l}_k$ is also one-to-one.
See the detailed analysis in \S E.2.3 in \citep{yun2020n}.

\paragraph{Preliminaries.}
As in \citep{yun2020n}, the upper bound for the unique id $\tilde{l}_i$ is:
\begin{align}
   \tilde{l}_i &:= l_i + \sum_{j=1}^{i-2} \delta^{-jd} (l_{i-j} - l_{i+1-j}) + \delta^{-(i-1)d}(l_1 - l_2) \nonumber\\
   &\leq l_i + \delta^{-d}\sum_{j=1}^{i-2}(l_{i-j}-l_{i+1-j}) + \delta^{-(i-1)d}(l_1 - l_2) \nonumber\\
   &= l_i + \delta^{-d}(l_2-l_i) + \delta^{-(i-1)d}(l_1 - l_2) \nonumber\\
   &=\delta^{-(i-1)d}l_1 - (\delta^{-(i-1)d}-\delta^{-d})l_2 - (\delta^{-d}-1)l_i\\
   &\leq \delta^{-(i-1)d} l_1 \leq \delta^{-(i-1)d}\left( (n-1)\Delta + \delta^{-d+1} -\delta \right)\\
   &\leq \delta^{-(i-1)d} (i-1+\delta)(\delta^{-d}-1) \leq n\delta^{-id} - \delta.\label{eq:upper_bound_selective_op}
\end{align}
Similarly, we have
\begin{align}
   l_n \leq n\delta^{-nd} - \delta.
\end{align}
Also, for any $n\geq1$, we have
\begin{align}\label{eq:ineq_for_allmax}
   \left( \frac{2n+1}{2n} \right) \leq \left( \frac{2n+1}{2n} \right)^2 \leq \cdots \leq \left( \frac{2n+1}{2n} \right)^n \leq 2
\end{align}

\paragraph{All-max-shift operations.} 
Following \citep{yun2020n}, to make the interval between $l_k$ are disjoint with each other, the all-max-shift operation $\Omega^l: \mathbb{R}^{d\times n} \to \mathbb{R}^{d\times n}$ is a self-attention layer defined as follows:
\begin{align}
   \Omega^l(\bm L; c) = \bm L + c\bm e^{(1)} \psi^l(\bm L; 0).
\end{align}
The $(1, k)$-th entry of $\Omega^l(\bm Z; c)$ reads
\begin{align}
   \Omega^l(\bm L; c)_{1,k} = L_{1,k} + c\psi^l(\bm L; 0)_k = L_{1,k}+c \max_{j\in \mathcal{A}^l_k}\bm u^T \bm L_j.\label{eq:all_max}
\end{align}
The main idea of all-max-shift operation is that, in the $i$-th layer, we will 'replace' the current 'column' by the maximum column within reach of sparse attention pattern $\mathcal{A}^i$.
In the next layer, the shifted max column will again be 'replaced' by the new maximum value within reach of the shifted column.
After $n$ steps or layers, all the first elements of each column will be replaced by the one in the maximum column, which is the dominated value.
The steps within the dominated element are greater than the intervals of the whole $l_n$.
So, for two different inputs $\bm L$, they $n$ entries are distinct, and the requirement \ref{def:contextual_mapping_rule2} in Definition~\ref{def:contextual_mapping} satisfied.

Without loss of generality, in contrast with the case of the cycle attention Eq.~\ref{eq:selective_attn} in the adaptive selective operation, the case of the stacked sampled attention is the same as in Fig.~\ref{fig:cycle}, with $l=1$.

\paragraph{First layer of all-max-shift.}
The input of the first all-max-shift operation is $\tilde{\bm L} \in \mathbb{R}^{d\times n}$.
Recall that $\bm u^T\tilde{\bm L} = [l_1, \tilde{l}_2, \tilde{l}_3, \dotso, \tilde{l}_n]$ and each element is $0<l_1<\tilde{l}_2<\tilde{l}_3<\dotso<\tilde{l}_n<n\delta^{-nd}-\delta$. The last inequality holds as in Eq.~\ref{eq:upper_bound_selective_op}.
Let the output of the first layers be $\bm M^1$.
The $k$-th element in the first row reads
\begin{align}
   M^1_{1,k} := \tilde{L}_{1,k} + 2n^2 \delta^{-nd-1} \max_{j\in \mathcal{A}^1_k}\bm u^T \tilde{\bm L}_j = \tilde{L}_{1,k} + 2n^2 \delta^{-nd-1} \bm u^T \tilde{\bm L}_{k+1 \mod n},
\end{align}
where with constant value $c=2n^2 \delta^{-nd-1}$ in Eq.~\ref{eq:all_max}, and for each column we will have
\begin{align}\label{eq:first_all_max_shift}
   \bm u^T \bm M^1_k = \bm u^T \tilde{\bm L}_k + 2n^2 \delta^{-nd-1} \bm u^T \tilde{\bm L}_{k+1\mod n},
\end{align}
as the first element of $\bm u$ is 1. Next, we see that $\bm u^T \bm M^1_k$ is dominated by the right term $2n^2 \delta^{-nd-1} \bm u^T \tilde{\bm L}_{k+1\mod n}$,
which is defined by for any $k, k' \in [n]$,
\begin{align}
   \bm u^T \tilde{\bm L}_{k+1\mod n} < \bm u^T \tilde{\bm L}_{k'+1\mod n} \Rightarrow \bm u^T \bm M_k < \bm u^T \bm M_{k'}.
\end{align}
This is because the minimum gap between $\bm u^T \tilde{\bm L}_{k+1}$ is $\delta$, and we have
\begin{align}
   \bm u^T \tilde{\bm L}_k < n\delta^{-nd} < 2n^2\delta^{-nd-1} \cdot \delta,
\end{align}
so if we have $\bm u^T \tilde{\bm L}_{k+1\mod n} < \bm u^T \tilde{\bm L}_{k'+1\mod n}$, it could determine the order $\bm u^T \bm M_k < \bm u^T \bm M_{k'}$, because $\bm u^T \tilde{\bm L}_k$ is within the minimum gap of the right term of Eq.~\ref{eq:first_all_max_shift}, and so cannot change the overall value.

\paragraph{Second layer of all-max-shift.}
As in the first layer, we define the output of this layer as $\bm M^2$, and the $k$-th element in the first row reads
\begin{align}
   M^2_{1,k} := M^1_{1,k} + 2n^2 \delta^{-nd-1} \max_{j\in \mathcal{A}^2_k}\bm u^T \bm M^2_{j} = M^1_{1,k} + 2n^2 \delta^{-nd-1} \bm u^T \bm M^2_{k+1\mod n},
\end{align}
so for each column, we have
\begin{align}
   \bm u^T \bm M^2_k &= \bm u^T \bm M^1_k + 2n^2 \delta^{-nd-1} \bm u^T \bm M^2_{k+1\mod n}\nonumber\\
   &= \bm u^T \tilde{\bm H}_k + 2n^2 \delta^{-nd-1} \bm u^T \tilde{\bm H}_{k+1\mod n} \nonumber\\
     & \quad + 2n^2 \delta^{-nd-1}(\bm u^T \tilde{\bm H}_{k+1\mod n} + 2n^2 \delta^{-nd-1} \bm u^T \tilde{\bm H}_{k+2\mod n})\nonumber\\
   &= \bm u^T \tilde{\bm H}_k + 4n^2 \delta^{-nd-1} \bm u^T \tilde{\bm H}_{k+1\mod n} + (2n^2 \delta^{-nd-1})^2 \bm u^T \tilde{\bm H}_{k+2\mod n}.
\end{align}
The last term domains $\bm u^T \bm M^2_k$, because the minimum gap of $\bm u^T \bm M^2_{k+1\mod n}$ is at least $\delta$, and
\begin{align}
   &\bm u^T \bm M^2_k - (2n^2 \delta^{-nd-1})^2 \bm u^T \tilde{\bm H}_{k+2\mod n} = \bm u^T \tilde{\bm H}_k + 4n^2 \delta^{-nd-1} \bm u^T \tilde{\bm H}_{k+1\mod n}\nonumber \\
   & <(1+4n^2\delta^{-nd-1})n\delta^{-nd} \leq (1+4n)n^2\delta^{-2nd-1} \leq (2n^2 \delta^{-nd-1})^2 \cdot \delta.
\end{align}
The last inequality holds due to
\begin{align}
   \left(\frac{1+2n}{2n}\right)^2 \leq 2 \Leftrightarrow 1+4n \leq 4n^2,
\end{align}
from Eq.~\ref{eq:ineq_for_allmax}.

\paragraph{Repeat all-max-shifts.}
After all $n$ layers we get $\bm M^n$, and $\bm u^T \bm M^n_k $ is dominated by
\begin{align}
   (2n^2\delta^{-nd-1})^n \max_{j\in \mathcal{A}^n_k}\bm u^T \tilde{\bm H}_j = (2n^2\delta^{-nd-1})^n \tilde{l}_n.
\end{align}
Because the remains in $\bm u^T\bm M^n_k$ have strictly upper-bound
\begin{align}
   \bm u^T \bm M^n_k - (2n^2\delta^{-nd-1})^n \tilde{l}_n &< \left( \sum_{i=0}^{n-1} 
   \begin{pmatrix}
      n\\
      i
   \end{pmatrix} (2n^2\delta^{-nd-1})^i \right)n\delta^{-nd} \\
   &\leq \left( \sum_{i=0}^{n-1} 
   \begin{pmatrix}
      n\\
      i
   \end{pmatrix} (2n)^i \right) (n\delta^{-nd-1})^{n-1} n\delta^{-nd}\\
   &=\left( (1+2n)^n - (2n)^n \right)(n\delta^{-nd-1})^n\cdot \delta \leq (2n^2\delta^{-nd-1})^n\cdot \delta.
\end{align}
The last inequality used $(1+2n)^n - (2n)^n \leq (2n)^n$ from Eq.~\ref{eq:ineq_for_allmax}.

\paragraph{Verifying Contextual Mapping.}
This matches the analysis in \S E.2.5 of \citep{yun2020n}.
As all $\bm u$ selective-shift operations and all-max operations are bijective, and $\bm u$ map each column (token) of the input to the unique id, the requirement~\ref{def:contextual_mapping_rule1} in the Definition~\ref{def:contextual_mapping} holds.
As $\bm u^T \bm M^n_k$ are all dominated by $(2n^2\delta^{-nd-1})\tilde{l}_n$, and different inputs $\bm L$ have different $\tilde{l}_n$ as $\tilde{l}_n$ is influenced by all $[l_1, l_2, \dotso, l_n]$, 
not all columns are the same for different inputs $\bm L$, and $\bm u^T$ is the unique mapping.
The interval may be written
\begin{align}
   \bm u^T \bm M^n_k \in [(2n^2\delta^{-nd-1})^n \tilde{l}_n, (2n^2\delta^{-nd-1})^n (\tilde{l}_n+\delta)].
\end{align}
The upper bound holds as other terms are less than $(2n^2\delta^{-nd-1})^n \cdot \delta$ in total (not the dominated term).
So as we can see the interval for all $\bm u^T \bm M^n_k$ are disjoint for different inputs, and the requirement~\ref{def:contextual_mapping_rule2} in the Definition~\ref{def:contextual_mapping} holds.


\end{document}